\documentclass[runningheads]{llncs}

\usepackage{eccv}

\usepackage{eccvabbrv}
\usepackage{adjustbox}
\usepackage{graphicx}
\usepackage{booktabs}
\usepackage{algorithm}
\usepackage{algpseudocode}
\usepackage{xcolor, colortbl}
\usepackage{multirow}
\usepackage{booktabs}
\usepackage{tabularx}
\usepackage{amssymb}
\usepackage{amsmath}
\usepackage{enumitem}
\usepackage[dvipsnames]{xcolor}
\usepackage{wrapfig}

\usepackage{blindtext}
\usepackage{subcaption}

\usepackage{tikz}
\newcommand{\tikzxmark}{%
\tikz[scale=0.23] {
    \draw[line width=0.7,line cap=round] (0,0) to [bend left=6] (1,1);
    \draw[line width=0.7,line cap=round] (0.2,0.95) to [bend right=3] (0.8,0.05);
}}

\newcommand{\task}[1]{{$\mathcal{US}^{(#1)}$}}
\newcommand{\tabletask}[1]{{$\boldsymbol{\mathcal{US}^{(#1)}}$}}

\usepackage{array}
\newcolumntype{H}{>{\setbox0=\hbox\bgroup}c<{\egroup}@{}}

\usepackage[accsupp]{axessibility}  

\usepackage{algorithm,algpseudocode}
\algnewcommand{\Initialize}[1]{%
  \State \textbf{Initialize:}
  \Statex \hspace*{\algorithmicindent}\parbox[t]{.8\linewidth}{\raggedright #1}
}

\algnewcommand{\Input}[1]{%
  \State \textbf{Input:}
  \Statex \hspace*{\algorithmicindent}\parbox[t]{.8\linewidth}{\raggedright #1}
}
\DeclareMathOperator*{\argmax}{argmax}
\DeclareMathOperator*{\argsort}{argsort}

\usepackage[colorlinks = true,
        urlcolor  = magenta]{hyperref}

\usepackage{orcidlink}

\begin{document}

\title{POET: \underline{P}rompt \underline{O}ffs\underline{e}t \underline{T}uning for Continual \\ Human Action Adaptation}


\author{Prachi Garg\inst{1} \and
K J Joseph\inst{3} \and
Vineeth N Balasubramanian\inst{3} \and 
Necati Cihan Camgoz\inst{2} \and
Chengde Wan\inst{2} \and
Kenrick Kin\inst{2} \and
Weiguang Si\inst{2} \and
Shugao Ma\inst{2} \and
Fernando De La Torre\inst{1}
}

\authorrunning{P. Garg et al.}

\institute{Carnegie Mellon University, USA \and
Meta Reality Labs \and
Indian Institute of Technology, Hyderabad}

\maketitle

\begin{abstract}
As extended reality (XR) is redefining how users interact with computing devices, research in human action recognition is gaining prominence. Typically, models deployed on immersive computing devices are static and limited to their default set of classes. The goal of our research is to provide users and developers with the capability to personalize their experience by adding new action classes to their device models continually.
Importantly, a user should be able to add new classes in a low-shot and efficient manner, while this process should not require storing or replaying any of user's sensitive training data. We formalize this problem as privacy-aware few-shot continual action recognition. Towards this end, we propose \textit{POET: \underline{P}rompt-\underline{o}ffs\underline{e}t \underline{T}uning}. While existing prompt tuning approaches have shown great promise for continual learning of image, text, and video modalities; they demand access to extensively pretrained transformers. Breaking away from this assumption, POET demonstrates the efficacy of prompt tuning a significantly lightweight backbone, pretrained exclusively on the base class data. We propose a novel spatio-temporal learnable prompt offset tuning approach, and are the first to apply such prompt tuning to \textit{Graph Neural Networks}. We contribute two new benchmarks for our new problem setting in human action recognition: (i) NTU RGB\texttt{+}D dataset for activity recognition, and (ii) SHREC-2017 dataset for hand gesture recognition. We find that POET consistently outperforms comprehensive benchmarks. \footnote{\footnotesize Source Code at \url{https://github.com/humansensinglab/POET-continual-action-recognition}}

\end{abstract}

\section{Introduction}
\label{sec:intro}

\begin{figure*}[t]
\centering
\includegraphics[width=0.99\textwidth]{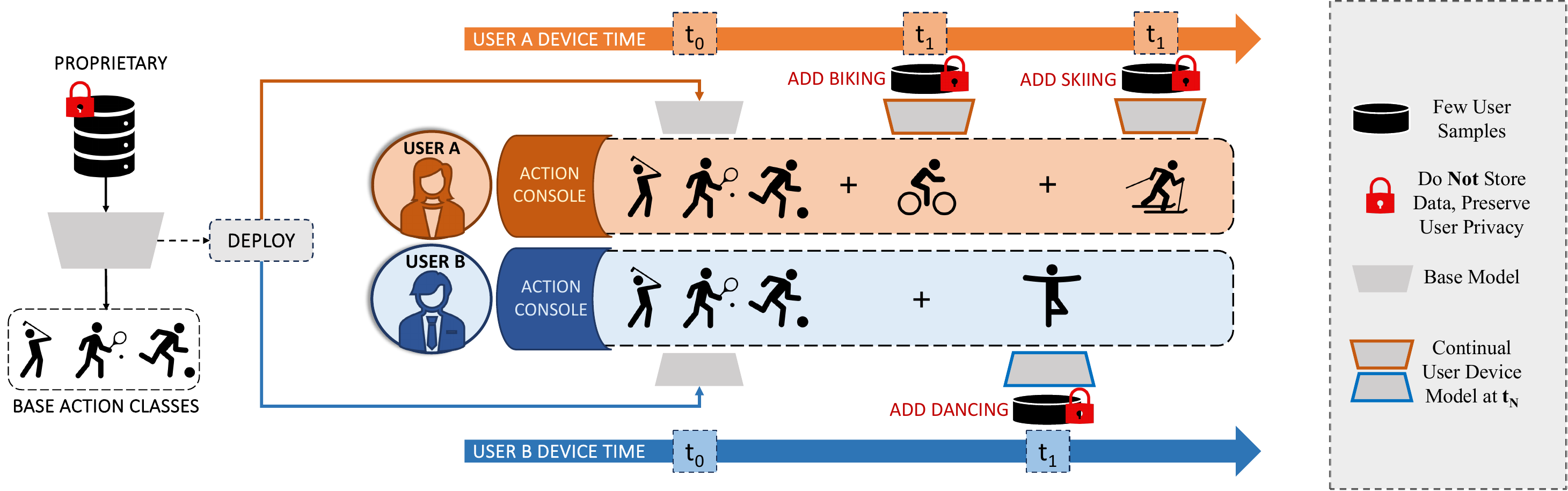}
\caption{Proposed POET method \textbf{continually adapts} skeleton-based human action recognition models pretrained on a pre-defined set of categories \textbf{to new user categories with few training examples}. Users can thus expand the capabilities of XR systems with novel action classes by providing a few examples of each new class. We discard the user-sensitive data as soon as the model is updated on the new categories.}
\vspace{-14pt}
\label{fig:teaser}
\end{figure*}

A key input modality to virtual, augmented and mixed reality (often together termed as extended reality, XR) devices today is through recognizing human activity and hand gestures based on body and hand pose estimates. Recognizing human actions\footnote{\scriptsize We use human action as an umbrella term for both hand gesture and body activity in this work for ease of presentation.} facilitates seamless user interactions in head-mounted XR devices such as the Meta Quest 3 and Apple Vision Pro.
If the provided action recognition models are static, then developers and users are limited to a predefined set of action categories.
With the growing use of such devices in new contexts and the increasing demand for personalized technology delivery, there is an impending need to enable the action recognition models in such systems to adapt and learn new user actions 
over time. Defining their own action categories allows users to customize their experience and expand the functionality of their XR devices. Addressing this need is the primary objective of this work.

Adapting human action models to new user categories over time faces a few challenges. Firstly, the model must be capable of learning new actions with minimal amount of training data so users can add new classes by providing just a few training examples per class. Secondly, due to the increasing use of XR devices for personal assistance, there is a need for privacy preservation in user action recognition-based pipelines \cite{hinojosa2022privhar, albrecht2016gdpr}. Hence, the adaptation of such action recognition models to new user categories must also be `data-free', i.e., it cannot store and replay previously seen user training data in subsequent continual sessions. Considering these requirements, we leverage the recent success of 
`data-free' prompt-based learning \cite{wang2022learning} and propose a new spatio-temporal prompt offset tuning approach to efficiently adapt the default model without finetuning.

Human action recognition systems are moving to skeleton-based approaches, especially in applications that require low-shot action recognition capabilities such as medical action recognition \cite{ma2022learning, zhu2023adaptive}. Skeletons offer a robust and compact alternative to videos in such low-shot regimes, due to their relatively low dimensionality and lesser variance under background conditions. While there have been a wide variety of efforts in skeleton-based human action recognition over the years \cite{yue2022action, ren2020survey, zhang2019comprehensive}, there have been fewer efforts on adapting such models to newer user categories. Efforts like \cite{li2021else,Aich2023Data} attempted to continually learn new user categories over time in skeleton-based human action recognition, but relied on fully-supervised data for the new classes. On the other hand, few-shot learning works \cite{ma2022learning, wang2023molo, zhu2023adaptive} adapt a pretrained skeleton-based action recognition model to new data, but without explicitly retaining past categories. In this work, we seek to learn new user categories in trained human action models with \textit{very few labeled samples} for the new classes, while being \textit{data-free} (not storing samples from previously trained categories). Fig \ref{fig:teaser} summarizes our overall objective. One could view our setting as privacy-aware few-shot continual learning for skeleton-based action recognition. 

To this end, we propose a prompt offset tuning methodology that can be integrated with existing backbone architectures for skeleton-based human action recognition. Our learnable (soft) prompts are selected from a shared knowledge pool of prompts based on an input instance dependent attention mechanism. In particular, we propose prompt selection using an ordered query-key matching that enables a temporal \textit{prompt} frame order selection consistent with the input instance. We show that such an approach allows us to learn new user categories without having to store data from past classes, without overwriting the pre-existing categories. To the best of our knowledge, this is the first effort on leveraging prompt tuning for skeleton-based models, as well as on spatio-temporal prompt selection and tuning.

\vspace{3pt}
\noindent Our key contributions are summarized below:
\vspace{-3pt}
\begin{itemize}[leftmargin=*]
\setlength\itemsep{-0.05em}
\item We formalize a novel problem setting 
which continually adapts human action models to new user categories over time, in a privacy aware manner. 
    \item 
    To address this problem, we propose a novel spatio-temporal Prompt OffsEt Tuning methodology (POET). In particular, it is designed to seamlessly plug-and-play with a pretrained model's input embedding, without any significant architectural changes.
    \item Our comprehensive experimental evaluation on two benchmark datasets brings out the efficacy of our proposed approach. 
    

\end{itemize}

\section{Related Works}
\label{sec_related_work}
\subsection{Prompt Tuning}
\label{sec: prompt tuning}
The idea of prompting, as it originated from Large Language Models (LLMs), is to include additional information, known as a text prompt, to condition the model's input for generating an output relevant to the prompt. Instead of applying a discrete, pre-defined `hard' language prompt token, \textit{prompt and prefix tuning} \cite{lester2021power, li2021prefix} formalized the concept of applying `soft prompts' to the input. A set of learnable parameters are prepended (concatenated) to the input text and trained along with the classifier while keeping the backbone parameters frozen. Similar to prompt tuning of LLMs, recent works have popularized prompt tuning of ViTs \cite{jia2022visual} as an effective way of adapting large pretrained models to downstream tasks \cite{wang2022learning, zhu2023prompt}. However, it remains unexplored and undefined (to the best of our knowledge) for \textit{non-transformer} architectures such as GNNs.
\subsection{Prompt Tuning for Continual Learning}
\label{sec: continual prompt tuning}
Prompt tuning provides a simple and cost-effective way of learning task-specific signal condensed into `soft prompts'. For continual learning, training a set of prompts for each sequential task provides a natural alternative to storing privacy violating exemplars and replaying them. Training task-specific prompts for each sequential task is straightforward when authors assume access to task identity at both train and inference time, like in Progressive Prompts \cite{razdaibiedina2023progressive}. However, if task identity is unavailable at inference, the model will not know which task's prompts or classifier to use for evaluating a test sample. In this respect, S-prompts \cite{wang2022s} and A-la-carte prompt tuning (APT) \cite{bowman2023carte} learn an independent set of prompts for each domain/task and employ a KNN-based search for domain/task identity at test time. Since these methods learn stand-alone prompts for every task, the prompt feature space is task-specific, and there is no forgetting of old knowledge when learning new tasks (by design). At the same time however, these `no forgetting' prompts \textit{cannot share knowledge} across tasks. 

This leads to another ideology for continual prompt tuning, i.e., treat each prompt unit as being a part of a larger \textbf{shared (knowledge) pool} of prompts. Then the desired number of prompt units can be selected from the pool, conditioned on the input instance itself \cite{wang2022learning, wang2022dualprompt, smith2023coda}. Given the scarcity of new data in our setting, we hypothesize that sharing of knowledge will benefit new tasks and draw inspiration from this line of works. Most recently, Adaptive Prompt Generator (APG) \cite{tang2023prompt} challenges the intensive ImageNet21K pre-training assumption as it prompts a ViT pretrained only on the continual benchmark's base class data (similar to us). However, they use 
replay and knowledge distillation-style `anti-forgetting learning', in addition to using prompts. Even though our backbone is trained only on the base classes, we propose a \textbf{simple prompt tuning-only} strategy to counter forgetting. This implies that a prompt strategy is all we need to continually add new action semantics in a few-shot manner. 
\subsection{Few-Shot Class Incremental Learning}
FSCIL is a challenging continual learning setting where a model overfits to new classes, with the simultaneous heightened (often complete) forgetting of old knowledge as soon as the base model is fine-tuned on few-shot data \cite{tao2020few, dong2021few}. Since the backbone feature extractor is the only source of previously seen knowledge, if it is updated, knowledge is lost forever. Typically, existing works decouple the learning of (backbone) feature representations from the classifier by \textit{learning} the model \textit{only on the base data} and relying on non-parametric class-mean classifiers for classification in subsequent steps \cite{peng2022few, zhou2022forward, hersche2022constrained}. This leads to a feature-classifier misalignment issue \cite{pernici2021class, yang2023neural} because new class prototypes are extracted from a backbone representation trained only on the base classes. We hypothesize that optimizing input prompt vectors along with a dynamically expanding parametric classifier on top of a frozen backbone can alleviate this misalignment issue. Our work not only provides a fresh perspective into FSCIL, but to our best knowledge is also the only work not designed for and evaluated on image benchmarks.

\vspace{-0.4cm}


\section{Preliminaries}
\vspace{-5pt}
\noindent \textbf{Skeleton Action Recognition Using Graph Representations.}
Our input $\mathbf{X} \in \mathbb{R}^{T \times J \times 3}$ is a video sequence of $T$ frames, each frame containing $J$ joints of the human body ($25$ joints) or hand skeleton ($22$ joints) in 3D Cartesian coordinate system. Such a skeleton action sequence is naturally represented as a graph topology $G = \{\mathcal{V}, \mathcal{E}\}$ with $\mathcal{V}$ vertices and $\mathcal{E}$ edges. Graphs are modeled using Graph Neural Networks (GNNs) \cite{dwivedi2022graph}, which can either be sparse graph convolutional networks (GCN) or fully connected graph transformers (GT). Our main model (a GNN) is defined as $f(\mathbf{X}) = f_c \circ f_g \circ f_e(\mathbf{X})$ (as also shown in Fig. \ref{fig: method, main model}). Input $\mathbf{X}$ is first passed through an input embedding layer $f_e$ to get an embedding of human joints $\mathbf{X_e} = f_e(\mathbf{X}), \mathbf{X_e} \in \mathbb{R}^{T \times J \times C_e}$, with feature dimension $C_e$. $\mathbf{X_e}$ is further passed to a graph feature extractor $f_g$ composed of a stack of convolutional layers (in GCNs) or attention layers (in GTs), and finally a classifier $f_c$ which predicts the action class label $\mathbf{y}$. In POET, we propose to attach learnable parameters $\mathbf{P_T}$ (called prompt offsets) to the embedding $\mathbf{X_e}$.

\vspace{4pt}
\noindent \textbf{Problem Definition.}
\label{method: problem def.}
Given a default (pre)trained model 
deployed on a user's device, we would like to extend this model to new action classes over $T$ subsequent user sessions (also called tasks) \{$\mathcal{US}^{(1)},..., \mathcal{US}^{(T)}$\}\footnote{\scriptsize User sessions may be spaced at arbitrary time intervals.}.
In each user session $\mathcal{US}^{(t)}$, the model learns a dataset $\mathcal{D}^{(t)} = {(\mathbf{X}_i^t, \mathbf{y}_i^t)}_{i=1}^{\mid \mathcal{D}^{(t)} \mid}$ of skeleton action sequence and label pairs provided by the user, $\mathbf{X}_i^t \in \mathbb{R}^{T \times J \times 3}$, $y_i^t \in \mathbb{R}^{\mathcal{Y}^{(t)}}$. In each session, the user typically provides a few training instances $F$ (e.g. $F \leq 5$) for each of the $N$ new classes being added, such that $|\mathcal{D}^{(t)}| = NF$. The base (default) model's session $\mathcal{UB}^{(0)}$ is assumed to have a large number of default action classes $\mathcal{Y}^{(0)}$ trained on sufficient data $\mathcal{D}^{(0)}$, which is most often proprietary and cannot be accessed in later user sessions. In each session, the user adds new action classes such that, $\mathcal{Y}^{(t)} \cap \mathcal{Y}^{(t')} = \emptyset, \forall t \neq t'$\footnote{\scriptsize We make this assumption considering this is a first of such efforts; allowing for overlapping action classes and users to `update' older classes would be interesting extensions of our proposed work.}. Due to the aforementioned privacy constraints, in any training session $\mathcal{US}^{(t)}$, the model has access to only $\mathcal{D}^{(t)}$; after training this data is made inaccessible for use in subsequent sessions (no exemplar or prototypes stored). After training on every new session $\mathcal{US}^{(t)}$, the model is evaluated on the test set of all classes seen so far $\cup_{i=0}^t \mathcal{Y}^{(i)}$.
The challenge is to alleviate forgetting of old classes while not overfitting to the user-provided new class samples. One could view our setting as privacy-aware few-shot continual action recognition, a problem of practical relevance in human action recognition -- which has not received adequate attention.


\vspace{-7pt}
\section{Methodology: Prompt Offset Tuning (POET)}
\label{sec: method}

\vspace{-5pt}
\noindent \textbf{Overview.} We propose to prompt tune a base GNN model $f(.)$ by prompts $\mathbf{P_T}$ to address our overall objective. As shown in Fig. \ref{fig: method, prompt selection}, for each input instance $\mathbf{X}$, corresponding prompts $\mathbf{P_T}$ are selected from a pool of prompt parameters, using an input-dependent query and key attention mechanism. The selected prompts are added to the input feature embedding (and hence the term `\textit{prompt offsets}'), before forwarding to the feature extractor and classifier (shown in Fig. \ref{fig: method, main model}).

\begin{wrapfigure}[24]{r}{0.35\textwidth}
\vspace{-10pt}
\includegraphics[width=0.34\textwidth]{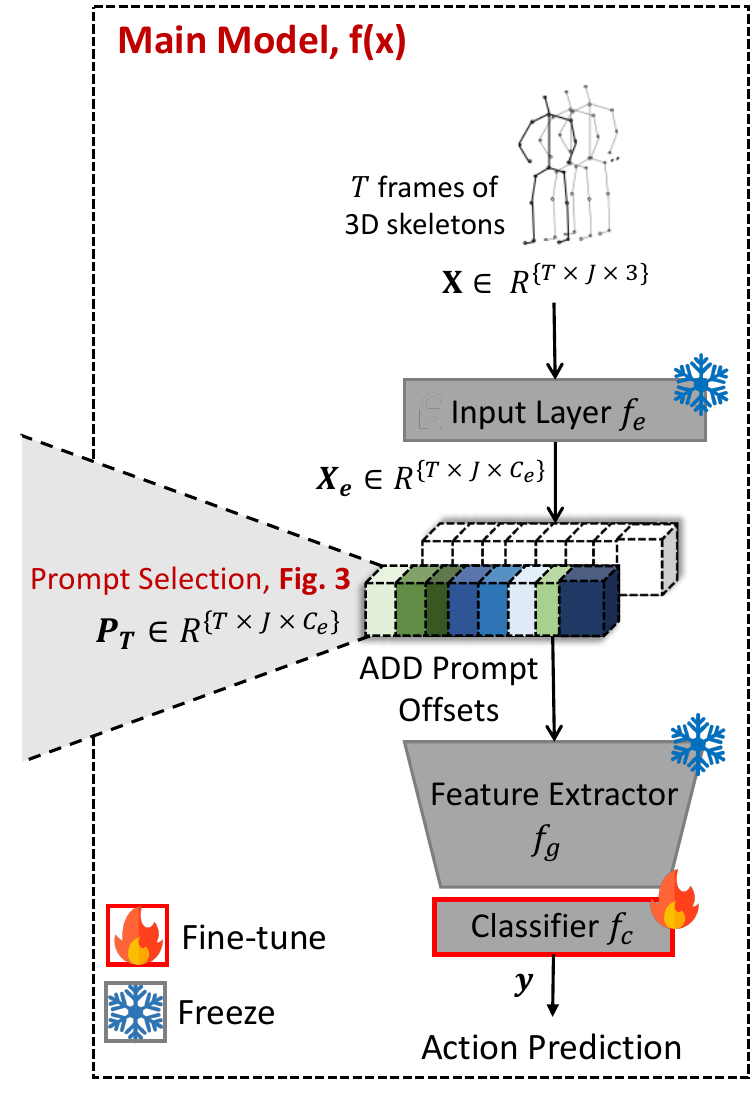}
\vspace{-10pt}
\caption{
\textbf{\scriptsize POET: \underline{P}rompt-\underline{o}ffs\underline{e}t \underline{T}uning} proposes to offset the input feature embedding $\mathbf{X_e}$ of the main model by learnable prompt parameters $\mathbf{P_{T}}$ for privacy-aware few-shot continual action recognition. 
We explain prompt selection mechanism in Fig. \ref{fig: method, prompt selection}.}
\label{fig: method, main model}
\end{wrapfigure}
To this end, our method, POET uses the same number of prompts as the number of temporal frames in the input, to maintain temporal consistency between the prompt and the input. Focusing solely on prompt offsets allows us to adapt the model to subsequent user sessions without having to update the input embeddings or the feature extraction backbone. Our prompt selection mechanism is learnable and trained along with the classifier to make this method simple and efficient. 


\vspace{4pt}
\noindent \textbf{What are Prompt Offsets?} Learnable (or soft \cite{lester2021power}) prompts are parameter vectors in a continuous space which are optimized to adapt the pretrained frozen backbone \textcolor{cyan}{$f_g$} to each continual task. We define our spatio-temporal prompt offsets $\mathbf{P_{T}}$ as a set of $T$ prompts (same in number as skeletal frames in input), each prompt $\boldsymbol{P_i}$ having length equal to the number of joints in a frame $J$ and feature dimension same as input feature embedding $\mathbf{X_e}$, i.e., $\boldsymbol{P_{i}} \in \mathbb{R}^{J \times C_e}$.

Existing prompt tuning efforts, for example in image classification, focus on concatenating learnable prompts to the input token sequence in transformer architectures \cite{lester2021power, jia2022visual}. Even though transformers can be generalized to graphs \cite{dwivedi2020generalization, mialon2021graphit, chen2019construct}, it is non-trivial to attach prompts to a GNN. This is because transformers can be viewed as treating sentences or images as fully connected graphs where any word (or image patch) can attend to any other word in the sentence \cite{dwivedi2022graph}. However, our input is a spatio-temporal graph skeleton of the human joint-bone structure with its own edge connectivity. Concatenating prompts along spatial or temporal dimensions would affect the graph semantics, and also affect standard training strategies such as a forward pass or backpropagation (especially in GCNs). Hence, we attach the selected prompts $\mathbf{P_T}$ to the corresponding input feature embedding $\mathbf{X_e}$ via a prompt attachment operator $f_p(.)$. The class logit distribution $\mathbf{y}$ is thus obtained as:
\vspace{-5pt}
\begin{equation} \label{eqn: 1}
    \mathbf{y} = f(\mathbf{X}, \mathbf{P_T}) = \textcolor{RedOrange}{f_c} \circ \textcolor{cyan}{f_g} \circ f_p(\textcolor{cyan}{f_e}(\mathbf{X}), \textcolor{RedOrange}{\mathbf{P_T}})
\vspace{-4pt}
\end{equation}

\noindent In every user session $t > 0$, the classifier output dimension expands by $N$ to accommodate the new action classes. Unlike most existing continual prompt tuning works, our feature extractor backbone $f_g$ is trained only on the base class data $\mathcal{D}^{(0)}$ and is never fine-tuned on classes from new user sessions $\mathcal{US}^{(t)}, t>0$. After the base session training, parameters of \textcolor{cyan}{$f_g, f_e$} are frozen. 
\begin{figure*}[t]
\centering
\includegraphics[width=0.95\textwidth]{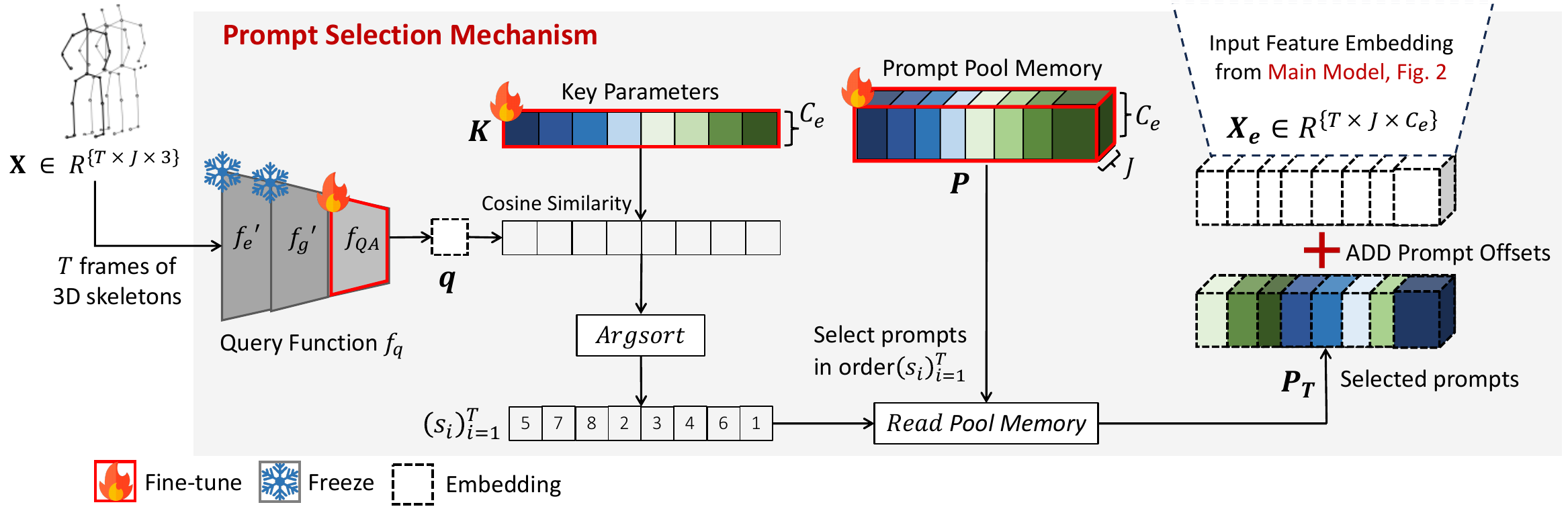}
\vspace{-0.15in}
\caption{\scriptsize \textbf{Selection of our prompts $\mathbf{P_{T}}$}: Input-dependent query $\boldsymbol{q}$ is matched with keys $\boldsymbol{K}$ using \textit{sorted} cosine similarity to get an ordered index sequence $(s_i)_{i=1}^{T}$ of the top $T$ keys. This ordered index sequence is used to select the corresponding ordered prompt sequence $\mathbf{P_{T}}$ from prompt pool $\mathbf{P}$. We \textit{add} $\mathbf{P_T}$ to $\mathbf{X_e}$, thereby adding an offset to it. Our experimental evaluation confirms that such an additive spatio-temporal prompt offset can balance the plasticity to learn new classes from a few action samples, while maintaining stability on previously learned classes.}
\vspace{-12pt}
\label{fig: method, prompt selection}
\end{figure*}

\vspace{4pt}
\noindent \textbf{Prompt Pool Design.} As stated in Sec.~\ref{sec: continual prompt tuning}, to encourage knowledge sharing across user sessions, we choose to construct a single prompt pool $\mathbf{P}$ which encodes knowledge across the sessions:
\vspace{-5pt}
\begin{equation} \label{eqn: 2}
    \mathbf{P} = \{\boldsymbol{P_1},.. \boldsymbol{P_i}, ..., \boldsymbol{P_M} \}, \qquad \boldsymbol{P_{i}} \in \mathbb{R}^{J \times C_e} ; M = \textrm{\#prompts at time t}
\vspace{-4pt}
\end{equation}

\noindent For selecting prompts from this pool (Fig. \ref{fig: method, prompt selection}), we construct a bijective key-value codebook, treating prompts in the pool $\mathbf{P}$ as values and defining learnable key vectors $\boldsymbol{K} = \{\boldsymbol{k_1},.., \boldsymbol{k_i},.., \boldsymbol{k_M} \}, \boldsymbol{k_i} \in \mathbb{R}^{C_e}$. A cosine similarity matching $\gamma(.)$ between the query $\boldsymbol{q}$ and keys $\boldsymbol{K}$ is used to find indices of the $T$ closest keys $\mathbb{Z}$, which in turn are used to select prompts from the pool: 
\vspace{-5pt}
\begin{equation} \label{eqn: 3}
 \mathbb{Z} = \argmax_{T} \gamma (f_q(\mathbf{X}), \boldsymbol{K})
\vspace{-3pt}
\end{equation}

\noindent This quantization process is enabled by a query function $f_q(.)$, which is a pre-trained encoder that maps an input instance $\mathbf{X}$ to a query $\boldsymbol{q}$ as:
\vspace{-5pt}
\begin{equation} \label{eqn: 4}
    \boldsymbol{q} = f_q(\mathbf{X}) = \textcolor{RedOrange}{f_{QA}} \circ \textcolor{cyan}{f_g'} \circ \textcolor{cyan}{f_e'} (\mathbf{X}), \qquad f_q \colon \mathbb{R}^{T \times J \times 3} \longrightarrow \mathbb{R}^{C_e}
\vspace{-3pt}
\end{equation}

\noindent where the \textit{query adaptor} \textcolor{RedOrange}{$f_{QA}$} is a fully connected layer mapping the \textcolor{cyan}{$f_g'$} output dimension to the desired prompt embedding dimension $C_e$. 

\vspace{4pt}
\noindent \textbf{Coupled Optimization in User Sessions $t>0$.} Typically, the $\argmax$ operator in Eq. \ref{eqn: 3} decouples the optimization of keys from the prompt pool and main model as it prevents backpropagation of gradients to the keys (also seen in earlier works such as \cite{wang2022learning, wang2022dualprompt}).
However, this approach does not work for our setting, even more so since we assume no large-scale pre-training of our base model. Due to the lack of off-the-shelf availability of large-scale models for skeletal action data, our query function $f_q$ is pretrained only on base class data $\mathcal{D}^{(0)}$. Hence, it becomes important that $f_q$ is updated as the model learns new classes. 
As shown in red boxes in Figs. \ref{fig: method, main model} and \ref{fig: method, prompt selection}, we propose to couple this optimization process such that the overall cross-entropy loss for new tasks updates: (i) the classifier \textcolor{RedOrange}{$f_c$}, (ii) selected prompts in \textcolor{RedOrange}{$\mathbf{P}$}, (iii) selected keys in \textcolor{RedOrange}{$\boldsymbol{K}$}, as well as (iv) query adaptor \textcolor{RedOrange}{$f_{QA}$}. We achieve this by approximating the gradient for \textcolor{RedOrange}{$\boldsymbol{K}$} and \textcolor{RedOrange}{$f_{QA}$} by the straight-through estimator reparameterization trick as in \cite{van2017neural, bengio2013estimating}. We freeze the query feature extractor layers \textcolor{cyan}{$f_g', f_e'$} in $t>0$ to prevent catastrophic forgetting of base knowledge in $f_q$. Our cross-entropy loss is hence given by:
\vspace{-5pt}
\begin{equation} \label{eqn: 5}
    \min_{\theta_{f_{QA}}, \theta_{\boldsymbol{K}}, \theta_{\mathbf{P}}, \theta_{f_c}} \mathcal{L}(f(\mathbf{X}, \mathbf{P_T}), \mathbf{y})
\vspace{-3pt}
\end{equation}

\noindent To move queries closer to their aligned $T$ keys during training, we use a vector quantization clustering loss inspired from VQ-VAE \cite{van2017neural} as:
\vspace{-5pt}
\begin{equation} \label{eqn: 6}
 \max_{\theta_{f_{QA}}, \theta_{\boldsymbol{K}}} \lambda \sum_{i \in \mathbb{Z}} \gamma (f_q(\mathbf{X}), \boldsymbol{K_i})
\vspace{-3pt}
\end{equation}

\noindent where $\lambda$ is the clustering loss coefficient. Our end-to-end optimization thus establishes a prompt optimization framework which is amenable to prompt tuning when extensive pre-training is not possible. This sets the foundation for our spatio-temporal prompt selection module, described next.

\begin{wrapfigure}[15]{r}{0.7\textwidth}
\vspace{-23pt}
\includegraphics[width=0.7\textwidth]{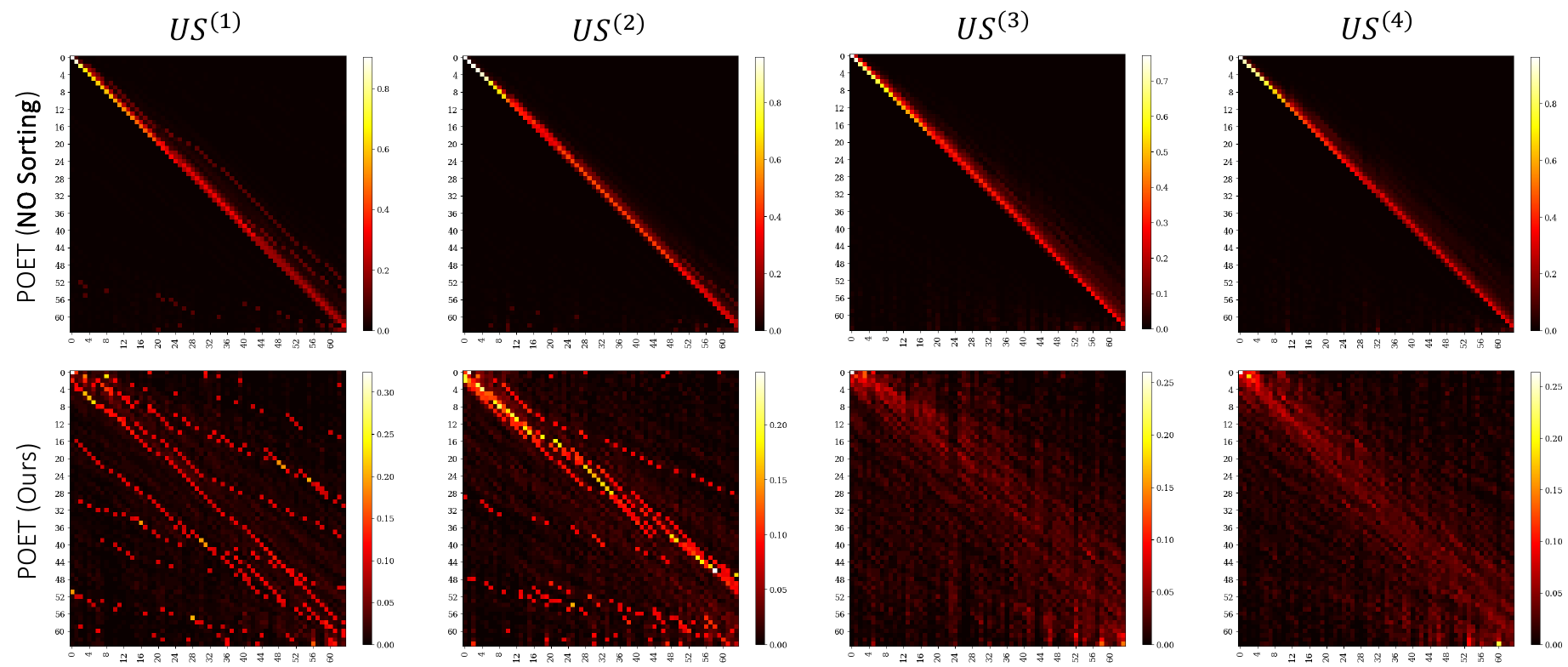}
\vspace{-14pt}
\caption{\scriptsize \textbf{Here we visualize the order $(s_i)_{i=1}^{T}$ in which} the $M = 64$ \textbf{prompts in the pool are selected} at train time, across 4 user sessions $\mathcal{US}^{(t)}$. X-axis: prompt index, Y-axis: index position in selected sequence. 
\textit{Top:} The no sorting case uses the default sequence (hence diagonal matrices), giving equal importance to all prompts.
\textit{Bottom (Our Method):} Even though the same 64 prompts are selected and updated, the ordering is temporally unique and consistent with input. }
\label{fig: ordering_qual_vis}
\end{wrapfigure}

\vspace{4pt}
\noindent \textbf{Spatio-Temporal Prompt Selection.} In order to ensure that our learned prompts respect temporal information in the input video sequence, we choose the number of selected prompts to be equal to the number of frames in the input 
\begin{wrapfigure}[13]{r}{0.5\textwidth}
\vspace{-23pt}
\includegraphics[width=0.5\textwidth]{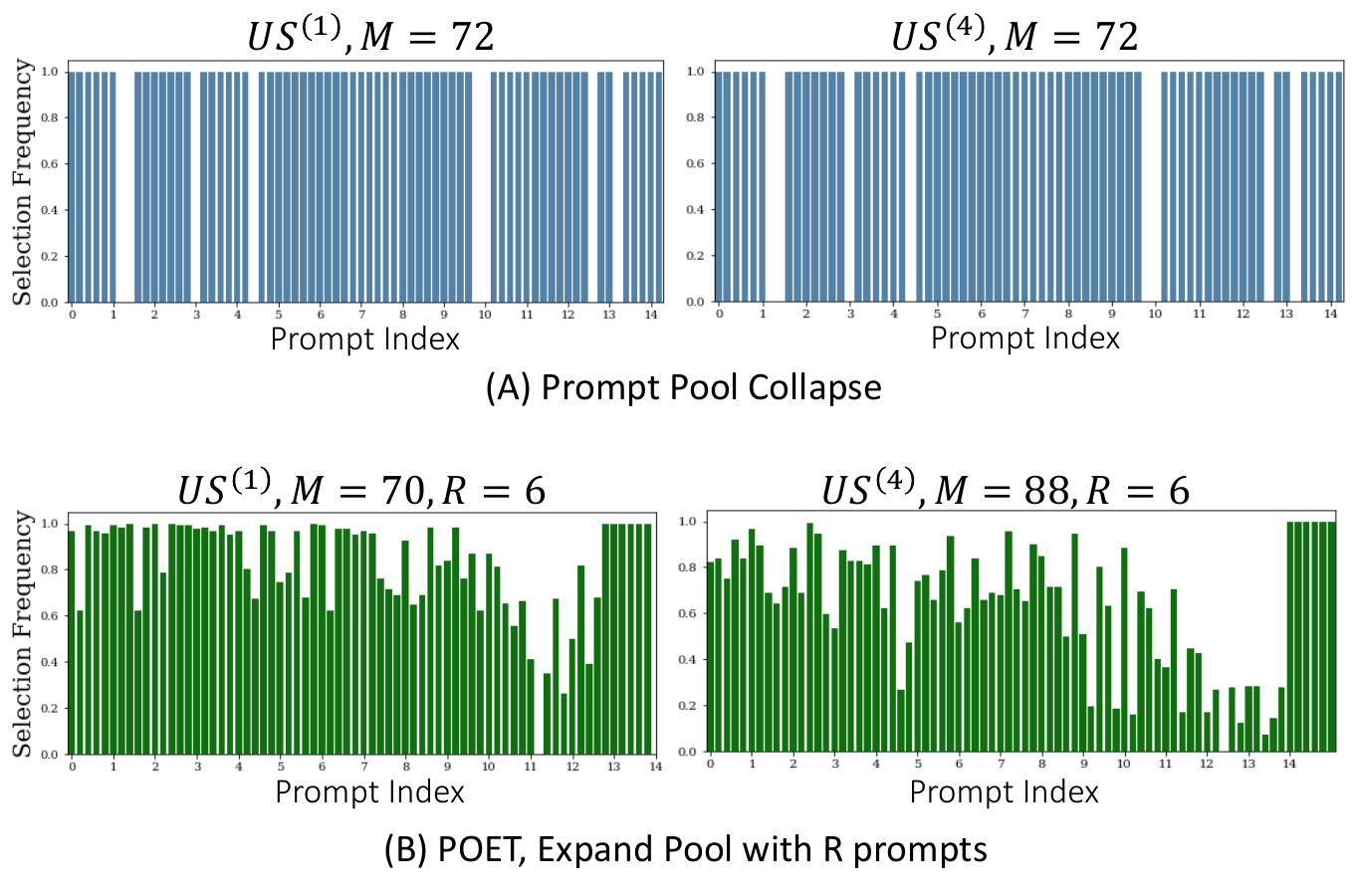}
\vspace{-22pt}
\caption{\scriptsize \textbf{$M>T$ Case:} \textbf{Prompt Pool Collapse.} (Top) Certain prompt indices remain unused across user sessions. (Bottom) Our POET pool expansion strategy alleviates pool collapse.}
\label{fig: pool collapse}
\end{wrapfigure}
video $T$. After coupling the prompt pool and keys, we observed in our initial experiments with pool size $M > T$ that the same set of prompts get selected across training iterations and user sessions (Fig. \ref{fig: pool collapse}A). More concretely, as the vector quantization loss (Eqn \ref{eqn: 6}) brings the query close to the selected keys, the same set of active prompts get selected and optimized in each iteration, not using other prompts at all. This is similar to the well-known issue of `\textit{codebook collapse}' in VQ-VAE \cite{williams2020hierarchical, dhariwal2020jukebox, zheng2023online}. Based on this observation, we design two prompt pool update mechanisms in user sessions $t>0$ as below: 
\vspace{-5pt}
\begin{enumerate}[leftmargin=*]
    \item \textbf{Case 1, $M = T \forall t$:} \textit{No pool expansion, Algorithm \ref{alg:POET}}. All prompts are selected in all tasks. But the \textit{order of their selection $(s_i)_{i=1}^{T}$} varies with each input instance as we replace Eq. \ref{eqn: 3} by sorting the cosine similarity before selecting the top $T$ indices as follows:
    \vspace{-5pt}
    \begin{equation} \label{eqn: 7}
     \mathbb{Z} = \argsort_{(s_i)_{i=1}^{T}} \gamma (f_q(\mathbf{X}), \boldsymbol{K})
    \vspace{-5pt}
    \end{equation}
    \noindent In Fig. \ref{fig: ordering_qual_vis}, we visualize the positions occupied by indices in this (sorted) \textit{ordered key index sequence} $(s_i)_{i=1}^{T}$. Entropy increase across tasks $t=1$ to $t=4$ (bottom row of figure) shows that our selection mechanism learns to select a unique temporal code for all inputs. 
    
    \item \textbf{Case 2, $M = T + (R * t), t > 0$.} \textit{Expand pool with R prompts}. We also propose an order-aware prompt pool expansion strategy (Appendix B) that selects prompts from an expanded pool in a temporally coherent manner, for $t>0$. This alleviates prompt pool collapse as shown in Fig. \ref{fig: pool collapse}B.  
\end{enumerate}

\noindent \textbf{Prompt Offset Attachment.} Since concatenation is not meaningful for graph data, we use addition as our choice for the prompt attachment operator as:
\vspace{-5pt}
\begin{equation} \label{eqn: 8}
 f_p(\mathbf{X_e}, \mathbf{P_T}) = \mathbf{X_e} + \mathbf{P_T}
\vspace{-3pt}
\end{equation}
Hence, we call our approach as \textit{prompt offset tuning}. We also study this empirically through experiments that support this choice in Sec. \ref{sec: ablations}. 

\vspace{4pt}
\noindent \textbf{Interpreting Prompt Offset Tuning of GNNs.} Our additive prompt offsets are open to interpretation, as shown in Fig. \ref{fig: ordering_qual_vis}. (i) Adding our selected prompts $\mathbf{P_T}$ to input feature embedding $\mathbf{X_e}$ acts like an input-dependent transformation 
for spatio-temporal joints. (ii) As our prompts have same size as $\mathbf{X_e}$, it can also be thought of as a learned prompt encoding, bearing similarity with learnable position encoding works \cite{liu2020learning, li2021learnable, dwivedi2022graph}. Our purpose is different however as prompt offsets seek to dynamically condition the input for adapting the backbone continually, instead of learning positions. (iii) POET also bears similarity with auto-decoders like DeepSDF \cite{park2019deepsdf} which learn latent codes for each style or shape and use relevant codes along with a frozen decoder at inference. (iv) Prompt tuning can also be thought of as a \textit{parameter isolation} technique for continual learning \cite{rebuffi2018efficient, mallya2018packnet, rusu2016progressive, razdaibiedina2023progressive}. POET's ordered prompt selection as seen in Fig. \ref{fig: ordering_qual_vis} learns to \textit{isolate} the relevant sequence of prompts for each input action sequence. 


\vspace{-0.4cm}
\begin{algorithm*}[bh!]
\scriptsize
\vspace{-2pt}
\caption{\footnotesize POET at Train Time, $t > 0$ (Case 1 $M=T$, No pool expansion)}\label{alg:POET}
\begin{algorithmic}
\State \textbf{Input:} Query function $f_q$, keys $\boldsymbol{K}=\{\boldsymbol{k_j}\}_{j=1}^T$, prompt pool $\mathbf{P}=\{\boldsymbol{P_j}\}_{j=1}^T$; main model $f_e$, $f_g$, $f_c$
\State \textbf{Initialize:} $\mathbf{P}, \boldsymbol{K}$ from $t-1$; Expand $f_c$ by $N$ new classes. Initialize $f_c$ as: (i) copy $f_c^{old}$ weights, (ii) $f_c^{new} \leftarrow Mean(f_c^{old})$ 
\State \textbf{Freeze:} query layers \textcolor{cyan}{$f_g', f_e'$}; main model layers \textcolor{cyan}{$f_e, f_g$}
\State \textbf{for} epochs and batch ${(\mathbf{X}_i^t, \mathbf{y}_i^t)}_{i=1}^{NK}$ do
    \State \hskip0.8em 1. Get query feature $\boldsymbol{q}$ (Eq. \ref{eqn: 4}) ; Compute $\gamma(.)$ b/w query $\boldsymbol{q}$ and keys $\boldsymbol{K}$
    \State \hskip0.8em 2. \textcolor{black}{Sort} $\gamma(.)$; Get \textcolor{black}{ordered key index sequence} $(s_i)_{i=1}^{T}$ (Eq. \ref{eqn: 7})
    \State \hskip0.8em 3. Read pool memory $\mathbf{P}$ in order $(s_i)_{i=1}^{T}$ $\rightarrow$ Get prompt offsets $\mathbf{P_T}$
    \State \hskip0.8em 4. Get $\mathbf{X_e}$; \textbf{Add} $\mathbf{P_T}$ to it (Eq. \ref{eqn: 8}); get prediction $\mathbf{y}$ from prompted input (Eq. \ref{eqn: 1})
    \State \hskip0.8em 5. Use cross entropy loss (Equation \ref{eqn: 5}) to \textbf{update} $f_{QA}, \boldsymbol{K}, \mathbf{P}, f_c$ 
    \State \hskip0.8em 6. Use clustering loss (Equation \ref{eqn: 6}) to \textbf{update} $f_{QA}$ and $\boldsymbol{K}$ 
\State \textbf{end} \textcolor{orange}{// See $t=0$ training protocol in Algorithm 2 in Appendix} 
\vspace{-2pt}
\end{algorithmic}
\end{algorithm*}

\vspace{-24pt}
\section{Experiments and Results}
\label{experiment section}
\vspace{-7pt}



\textbf{Datasets.} We evaluated our method on well-known action recognition datasets\footnote{The datasets used in this work were accessed and processed at and by CMU. They were not accessed, processed, stored, or maintained at Meta.}: (i) activity recognition on the NTU RGB+D dataset \cite{shahroudy2016ntu}; and (ii) hand gesture recognition on the SHREC-2017 dataset \cite{10.2312:3dor.20171049}. As we introduce a new problem setting in human action recognition, we contribute two new benchmarks to the community for this setting, on the NTU RGB+D and SHREC-2017 datasets. 

For the NTU RGB+D dataset, we divide the 60 daily action categories into 40 base classes, learning the remaining 20 classes in subsequent user sessions. In few-shot learning parlance, our protocol is 4-task 5-way 5-shot, i.e. 5 novel classes using 5 user training instances in 4 user sessions. 
Each input 3D skeleton sequence has $64$ temporal frames, each consisting of $25$ body keypoints, such that $x \in \mathcal{R}^{64 \times 25 \times 3}$. We use the spatio-temporal GCN, CTR-GCN \cite{chen2021channel}, as the architecture for NTU RGB+D, where we choose the joint input modality for better interpretability of prompt tuning. 

For SHREC-2017, we divide the 14 fine-grained hand gesture classes into 8 base classes and 6 classes learned in subsequent user sessions. This is done in a 3-task 2-way 5-shot protocol, i.e. 2 novel classes using 5 user training instances in 3 user sessions. 
For each input instance of SHREC-2017, we use $8$ temporal frames each having $22$ hand keypoints, such that input $x \in \mathcal{R}^{8 \times 22 \times 3}$. We use a fully-connected graph transformer backbone, DG-STA \cite{chen2019construct} for SHREC-2017. We select DG-STA due to easily reproducible code and to validate if our method POET works equally well across graph convolutional networks and graph transformers.

\noindent \textbf{Evaluation Metrics.}
Following earlier work in similar settings \cite{peng2022few}, we report: (i) Average accuracy `\textit{Avg}' of all classes seen so far, and (ii) Harmonic Mean $A_{HM}$ between `\textit{accuracy only on Old classes}' and `\textit{accuracy only on New classes}' after learning each new user session. Note that the average accuracy tends to be biased towards the base session $\mathcal{T}^{(0)}$ performance due to more number of base classes. A higher $A_{HM}$ implies better stability-plasticity trade-off between new task performance and old tasks' retention. 
Unlike many earlier CIL efforts, we report accuracy for both \textit{Old} and \textit{New} classes in each user session for transparency. 

\noindent \textbf{Implementation Details.} 
We observe that a key source of forgetting in our setting is from the classifier as the logits tend to become heavily biased towards the few-shot samples of new classes. We use a cosine classifier for activity recognition experiments on CTR-GCN. For gesture recognition on the lightweight DG-STA, we use a standard fully-connected layer as classifier, but freeze old class parameters in the classifier by zeroing their gradients. We attach prompts after the 1st layer of DG-STA and 1st CTR-GC block of CTR-GCN. 
For both datasets, we have equal or higher learning rates in user sessions when compared to the base model's training in order to accommodate new knowledge in the model (for better plasticity). For exact implementation details (including learning rates, epochs, hyperparameter analysis, and backward forgetting metric), see Appendix A. In earlier efforts that more generally tune prompts for class-incremental learning \cite{wang2022learning, wang2022dualprompt, smith2023coda, wang2022s, villa2023pivot}, it is common to rely on an ImageNet21K pretrained ViT \cite{ridnik2021imagenet} or CLIP \cite{radford2021learning} as the backbone. However, such backbones do not exist for skeleton-based human action recognition. Our base feature extractor is hence trained on the base session dataset itself without \textit{any} pretraining, making this one of the first efforts of prompt tuning without extensive pretraining (scale of data 3-5 times lower order of magnitude). 
\vspace{4pt}
\noindent \textbf{Results.} Since there are no existing baselines for our proposed setting in skeletal action recognition, we compare our method by adapting continual learning (CL) baselines to skeletal data in Sec \ref{sec: SOTA}, Tables \ref{sota activity}, \ref{SOTA gesture}. We first compare POET with prompt tuning based class-incremental learning (CIL) approaches originally designed for images (L2P \cite{wang2022learning}, CODA-P \cite{smith2023coda}, APT \cite{bowman2023carte}) and find that it has very low performance on new classes as they do not update their query function. We find any fine-tuning or knowledge distillation based approaches (LWF \cite{li2017learning}, EWC\cite{kirkpatrick2017overcoming}, LUCIR \cite{hou2019learning}) lead to rapid forgetting of base knowledge as the model overfits to user's few-shots. We also compare with multiple variants of Feature Extraction (FE) to check if prompts truly have merit (POET=FE+Prompts) and provide upper bound baselines. In Sec \ref{sec: ablations}, we first show the importance of prompts in POET by removing the prompts. We discuss the value of our coupled optimization, query function update and \textit{ordered key index selection} in our prompt selection ablation Tab \ref{tab: main ablation}. We also study the impact of proposed additive prompt tuning as compared to other possible prompt attachments $f_p$ in Tab \ref{tab: prompt attachment}.  

\subsection{Comparison with State-of-the-Art}
\label{sec: SOTA}

\noindent \textbf{POET sets the SOTA on existing prompt tuning works (Tab \ref{sota activity},\ref{SOTA gesture}).}  
We adapt three standard CIL works that prompt tune ViTs for images - L2P \cite{wang2022learning}, CODA-P \cite{smith2023coda} and APT \cite{bowman2023carte} to our setting. L2P and CODA-P share prompt pool across tasks (similar to us), whereas APT learns task-specific prompts. L2P decouples the optimization of keys from the prompt pool and concatenates the selected prompts. Since concatenation is not defined for our GNN backbone, we adapt these SOTA to our setting by concatenating along the temporal dimension (\textbf{L2P*, CODA-P*}).
CODA-P \cite{smith2023coda} couples keys with the prompt pool by using a cosine similarity weighing over all prompts in the pool, forming a `soft prompt selection', different from our `ordered hard prompt selection'. In \textbf{APT}, we train prompt-classifier pairs for each continual task separately (\^\ denotes task-specific), and use task identity at test time. See details in Appendix A. These methods by design rely on extensively pretrained (ImageNet21k) 
query functions which does not require updates; and require full supervision on new classes, perhaps explaining their poor `New' accuracy in our few-shot setting.

\noindent \textbf{Standard Continual learning Baselines.} We compared with two well established knowledge-distillation approaches, learning without forgetting (\textbf{LWF}) and \textbf{LUCIR}. Both of them perform poorly on both old and new classes. \textbf{EWC} \cite{kirkpatrick2017overcoming} learns better on new but does not retain old knowledge. We conclude that any CL method that fine-tunes the backbone feature representation in subsequent sessions $t>0$ will not be able to retain base/old class knowledge (a finding consistent with existing FSCIL literature for images \cite{dong2021few, tao2020few}). We also adapt and compare with one of the latest FSCIL baselines \textbf{ALICE} \cite{peng2022few}, originally developed for image classification benchmarks on our gesture recognition benchmark in Table \ref{SOTA gesture}. Note the high retention of base task performance (due to non-parametric classifier on top of frozen base model). However, it suffers from poor plasticity and adaptation to new classes. This is the issue of feature-classifier misalignment that we hoped to alleviate through prompt tuning. 


\begin{table*}[ht]
\setlength{\tabcolsep}{2.5pt}
\centering
\caption{\scriptsize \textbf{Activity Recognition Results (\%, $\uparrow$), Comparison with SOTA:} NTU RGB+D \cite{shahroudy2016ntu} dataset on CTR-GCN \cite{chen2021channel} backbone. After training on each incremental task, we report Average of all classes seen so far (`Avg'). We also report (i) $A_{HM}$, (ii) old classes accuracy (`Old'), (iii) new classes accuracy (`New') in the last session. We report Mean and STD across 10 sets of 5-shots. \textit{POET achieves the best stability-plasticity trade-off across all baselines indicated by the $A_{HM}=56.3\%$. POET also has the highest Avg across all user sessions outside of upper bound baselines \textcolor{orange}{in orange}.}}
\vspace{-6pt}
\begin{adjustbox}{width=0.96\textwidth}
\begin{tabular}{@{}l||c|HHc|HHc|HHc|cccc@{}} \toprule 
& \normalsize $\boldsymbol{\mathcal{UB}^{(0)}}$
& \multicolumn{3}{c}{\normalsize \tabletask{1}} 
& \multicolumn{3}{c}{\normalsize \tabletask{2}}
& \multicolumn{3}{c}{\normalsize \tabletask{3}}
& \multicolumn{4}{c}{\normalsize \tabletask{4}} \\
\cmidrule(lr){2-2} \cmidrule(lr){3-5} \cmidrule(lr){6-8} \cmidrule(lr){9-11}  \cmidrule(l){12-15} 
\textbf{Method} & Base ($\uparrow$) & Old ($\uparrow$) & New ($\uparrow$) & Avg ($\uparrow$) & Old ($\uparrow$) & New ($\uparrow$) & Avg ($\uparrow$) & Old ($\uparrow$) & New ($\uparrow$) & Avg ($\uparrow$) & Old ($\uparrow$) & New ($\uparrow$) & Avg ($\uparrow$) & $A_{HM}$ ($\uparrow$) \\
\hline
\hline
\textit{Upper Bounds}   \\

\cellcolor{orange!15}Joint (Oracle)  & 88.4 &  &  & 79.0 &  &  & 71.0 &  &  & 66.8 &  &  & \cellcolor{gray!20}\textbf{63.5} &  \\
\cellcolor{orange!15}Joint POET (Oracle) &  &  &  &  &  &  &  &  &  &  & & & \cellcolor{gray!20}\textbf{67.2} &  \\
\cellcolor{orange!15}FE, Task-Specific\^ & 88.4 &  &  & 70.1 $\pm$ \scriptsize 2.6 &  &  & 52.5 $\pm$ \scriptsize 5.8 &  &  & 44.8 $\pm$ \scriptsize 5.0 & 70.3 $\pm$ \scriptsize 2.1  &  46.7 $\pm$ \scriptsize 2.0 & NA & NA \\

\cellcolor{orange!15}FE$+$Replay & 88.4 &  &  & 82.4 $\pm$ \scriptsize 1.1 &  &  & 78.2 $\pm$ \scriptsize 1.2 &  &  & 74.5 $\pm$ \scriptsize 1.2 & \textbf{73.1 $\pm$ \scriptsize 1.0} & 43.3 $\pm$ \scriptsize 3.3 & \textbf{70.6 $\pm$ \scriptsize 1.2} & 54.3 $\pm$ \scriptsize 2.6  \\
\\[-1.9ex]
\hline
\textit{Continual Linear Probing}   \\
\cellcolor{gray!15}FE & \cellcolor{gray!15}88.4 &  &  & \cellcolor{gray!15}72.0 $\pm$ \scriptsize 1.1 &  &  & \cellcolor{gray!15}60.4 $\pm$ \scriptsize 2.4 &  &  & \cellcolor{gray!15}47.7 $\pm$ \scriptsize 2.1 & \cellcolor{gray!15}40.0 $\pm$ \scriptsize 1.6 & \cellcolor{gray!15}51.0 $\pm$ \scriptsize 2.3 & \cellcolor{gray!15}40.9 $\pm$ \scriptsize 1.4 & \cellcolor{gray!15}44.8 $\pm$ \scriptsize 1.1 \\
FE, Frozen & 88.4 &  &  & 76.1 $\pm$ \scriptsize 1.0 &  &  & 52.4 $\pm$ \scriptsize 4.1 &  &  & 38.3 $\pm$ \scriptsize 2.7 & \cellcolor{white!20}28.4 $\pm$ \scriptsize 1.6 & \cellcolor{white!30}22.4 $\pm$ \scriptsize 4.5 & 27.9 $\pm$ \scriptsize 1.4 & 24.8 $\pm$ \scriptsize 3.0 \\

FE$+$Replay$\dagger$ & 88.4 &  &  &  72.0 $\pm$ \scriptsize 1.5 &  &  & 59.5 $\pm$ \scriptsize 4.0 &  &  & 58.7 $\pm$ \scriptsize 2.8 & 56.7 $\pm$ \scriptsize 2.5 & 34.7 $\pm$ \scriptsize 5.6 & 54.9 $\pm$ 2.7 \scriptsize & 42.8 $\pm$ \scriptsize 4.4 \\

FT & 88.4 & \phantom{0} &  & \phantom{0}6.2 $\pm$ \scriptsize 1.4 & \phantom{0} &  & \phantom{0}4.3 $\pm$ \scriptsize 1.5 & \phantom{0} &  & \phantom{0}2.8 $\pm$ \scriptsize 1.0 & \cellcolor{white!20}\phantom{0}0.2 $\pm$ \scriptsize 0.5 & \cellcolor{white!30}36.0 $\pm$ \scriptsize 10.1 & \phantom{0}3.2 $\pm$ \scriptsize 0.8 & \phantom{0}0.3 $\pm$ \scriptsize 1.0 \\
\\[-1.9ex]
\hline
\textit{Standard Continual Learning}   \\
LWF \cite{li2017learning} & 88.4 &  &  & \phantom{0}6.2 $\pm$ \scriptsize 1.5  &  & & \phantom{0}2.8 $\pm$ \scriptsize 0.7 &  & & \phantom{0}3.7 $\pm$ \scriptsize 1.3 & \phantom{0}0.0 $\pm$ \scriptsize 0.0 & 38.9 $\pm$ \scriptsize 8.8 & \phantom{0}3.2 $\pm$ \scriptsize 0.7 & \phantom{0}0.0 $\pm$ \scriptsize 0.0 \\
EWC \cite{kirkpatrick2017overcoming} & 88.4 & \phantom{0} & & \phantom{0}6.6 $\pm$ \scriptsize 1.5 & \phantom{0} &  & \phantom{0}4.1 $\pm$ \scriptsize 1.4 & \phantom{0} &  & \phantom{0}3.1 $\pm$ \scriptsize 0.9 & \phantom{0}0.0 $\pm$ \scriptsize 0.0 & 42.1 $\pm$ \scriptsize 9.5 & \phantom{0}3.5 $\pm$ \scriptsize 0.8 & \phantom{0}0.0 $\pm$ \scriptsize 0.0 \\
Experience Replay & 88.4 &  &  & 35.1 $\pm$ \scriptsize 8.3 &  &  & 50.6 $\pm$ \scriptsize 5.0 &  &  & 60.6 $\pm$ \scriptsize 5.4 & 54.6 $\pm$ \scriptsize 6.5  & 43.7 $\pm$ \scriptsize 14.6 & 53.7 $\pm$ \scriptsize 7.1 & 47.8 $\pm$ \scriptsize 11.2 \\
Experience Replay$\dagger$ & 88.4 &  &  & \phantom{0}6.2 $\pm$ \scriptsize 1.5 &  & & \phantom{0}9.0 $\pm$ \scriptsize 2.6 &  &  & 11.2 $\pm$ \scriptsize 3.0 & 10.9 $\pm$ \scriptsize 2.6 & 34.6 $\pm$ \scriptsize 7.9 & 12.9 $\pm$ \scriptsize 3.0 & 16.3 $\pm$ \scriptsize 3.5 \\
LUCIR \cite{hou2019learning} & 87.9 & \phantom{0}0.0 &  & \phantom{0}4.3 $\pm$ \scriptsize 2.1 & \phantom{0}0.0 &  & \phantom{0}4.1 $\pm$ \scriptsize 1.3 & \phantom{0}0.0 & & \phantom{0}2.7 $\pm$ \scriptsize 0.8 & \phantom{0}0.2 $\pm$ \scriptsize 0.4 & 26.0 $\pm$ \scriptsize 9.2 & \phantom{0}2.3 $\pm$ \scriptsize 0.9 & \phantom{0}0.4 $\pm$ \scriptsize 0.8 \\

\\[-1.9ex]
\hline
\textit{Continual Prompt Tuning}  \\ 
\cellcolor{white!15}CODA-P \cite{smith2023coda}* & \cellcolor{white!15}87.4 &  &  & \cellcolor{white!15}76.1 $\pm$ \scriptsize 1.0 & &  & \cellcolor{white!15}66.7 $\pm$ \scriptsize 1.3 &  &  & \cellcolor{white!15}58.6 $\pm$ \scriptsize 2.7 & \cellcolor{white!15}56.5  $\pm$ \scriptsize 2.9  & \cellcolor{white!15}\phantom{0}0.5  $\pm$ \scriptsize 0.4 & \cellcolor{white!15}51.8  $\pm$ \scriptsize 2.7  & \cellcolor{white!15}\phantom{0}1.1  $\pm$ \scriptsize 0.7  \\


\cellcolor{white!15}L2P \cite{wang2022learning}* & \cellcolor{white!15}88.6 &  &  & \cellcolor{white!15}78.9 $\pm$ \scriptsize 0.1  &  &  & \cellcolor{white!15}71.0 $\pm$ \scriptsize 1.0 &  &  & \cellcolor{white!15}64.2 $\pm$ \scriptsize 0.1 & \cellcolor{white!15}62.0 $\pm$ \scriptsize 0.7 & \cellcolor{white!15}\phantom{0}0.0 $\pm$ \scriptsize 0.0 & \cellcolor{white!15}56.8 $\pm$ \scriptsize 0.6  & \cellcolor{white!15}\phantom{0}0.0 $\pm$ \scriptsize 0.0  \\


APT \cite{bowman2023carte}\^ & 86.6 &  &  & 27.3 $\pm$ \scriptsize 1.6 &  &  & 30.8 $\pm$ \scriptsize 3.4 & \phantom{0}NA &  & 37.6 $\pm$ \scriptsize 2.3 & \phantom{0}NA & 33.4 $\pm$ \scriptsize 2.0 & \phantom{0}NA & \phantom{0}NA \\
\\[-1.9ex]
\hline
\cellcolor{gray!20}POET (Ours) & \cellcolor{gray!20}87.9 & \cellcolor{gray!20} & \cellcolor{white!30} & \cellcolor{gray!20}\textbf{82.3 $\pm$ \scriptsize 0.6} & \cellcolor{white!20} & \cellcolor{white!30} & \cellcolor{gray!20}\textbf{76.8 $\pm$ \scriptsize 0.9} & \cellcolor{gray!20} & \cellcolor{gray!20} & \cellcolor{gray!20}\textbf{68.4 $\pm$ \scriptsize 0.7} & \cellcolor{gray!20}\textbf{57.2 $\pm$ \scriptsize 1.0} & \cellcolor{gray!20}\textbf{55.8 $\pm$ \scriptsize 5.9} & \cellcolor{gray!20}\textbf{57.1 $\pm$ \scriptsize 1.1} & 
\cellcolor{gray!20}\textbf{56.3 $\pm$ \scriptsize 3.2} \\

\bottomrule
\end{tabular}
\end{adjustbox}
\label{sota activity}
\vspace{-10pt}
\end{table*}





\noindent \textbf{Fine-tuning (FE) and Feature Extraction (FE) Baselines.} 
We implement standard continual learning baselines to understand stability-plasticity trade-offs in our new benchmarks. In all these baselines, we expand the classifier output dimension by $N$ new classes. In \textbf{`FT (Fine-Tuning)'}, we tune all model parameters on cross entropy loss of new task. FSCIL is challenging for this modality as old task performance sharply reduces to zero starting from \task{1} as model overfits to user's few-shots. \textbf{`FE (Feature Extraction)'}\footnote{\scriptsize `FE' is the same as `w/o prompts' in Table \ref{tab: main ablation}. We highlight key baselines in gray color.} differs from FT as we freeze the feature extractor to preserve base knowledge. This serves as a competitive baseline in our findings. In \textbf{`FE, frozen'}, we zero out the gradients of previous class weights in classifier $f_c$ to prevent forgetting from the classifier. `FE' and `FE, Frozen' exhibit different New-Old trade-offs in Tables \ref{sota activity}, \ref{SOTA gesture} because the scale of pretraining is different (gesture more lightweight than activity).

\noindent \textbf{Upper-bound baselines, top section Tables \ref{sota activity}, \ref{SOTA gesture}.} 
In \textbf{`Joint (oracle)'} experiment, we train on all task data at the same time in a multi-task (non-sequential) manner. Training POET in a multi-task manner (\textbf{`Joint POET'}) outperforms \textbf{`Joint Oracle'} demonstrating the strength of our approach. In addition to these \textit{generalist} upper bounds, we point out that `\textbf{FE, Task-specific\^}' is a competitive \textit{specialist} upper bound. In this, we perform feature extraction from base model to each task individually, storing separate task-specific models (\task{0} $\rightarrow$ \task{i}, $i>0$). POET outperforms `New' accuracy compared with this baseline, achieving a forward transfer on each $t>0$. This indicates that prompt tuning benefits New performance due to the pre-existing knowledge in the shared knowledge pool. Avg in sessions $0<t<4$ indicates New for task-specific\^\ .

\noindent \textbf{Experience Replay Baselines, Tab \ref{sota activity}.} Even though our privacy-aware setting prohibits previous data replay, we compare with \textbf{`Experience Replay'} (store and replay 5-samples of base and incremental sessions) and \textbf{`Experience Replay$\dagger$'} (replay only previous incremental sessions) for completeness. \textbf{`FE+Replay'} serves as the best upper bound (even better than Experience Replay as we are freezing backbone in addition to replay). It is noteworthy that POET (which is FE+prompts) learns an implicit `data-free' form of prompt pool memory, and yet has a better $A_{HM}$ trade-off as compared to explicitly stored and replayed samples from previous classes in FE+replay.   

\begin{table*}[t]
\setlength{\tabcolsep}{3pt}
\centering
\caption{\scriptsize \textbf{Gesture Recognition Results (\%, $\uparrow$), Comparison with SOTA:} SHREC 2017 \cite{10.2312:3dor.20171049} dataset on DG-STA \cite{chen2019construct} graph transformer backbone. Reporting mean and standard deviation across 5 runs. \textbf{POET achieves best $A_{HM}=56.2\%$}.}
\label{SOTA gesture}
\vspace{-6pt}
\begin{adjustbox}{width=0.85\textwidth}
\begin{tabular}{@{}l||c|HHc|HHc|cccc@{}} \toprule 
& \normalsize $\boldsymbol{\mathcal{UB}^{(0)}}$
& \multicolumn{3}{c}{\normalsize \tabletask{1}} 
& \multicolumn{3}{c}{\normalsize \tabletask{2}}
& \multicolumn{3}{c}{\normalsize \tabletask{3}} \\
\cmidrule(lr){2-2} \cmidrule(lr){3-5} \cmidrule(lr){6-8} \cmidrule(l){9-12} 
\textbf{Method} & Base ($\uparrow$) & Old ($\uparrow$) & New ($\uparrow$) & Avg ($\uparrow$) & Old ($\uparrow$) & New ($\uparrow$) & Avg ($\uparrow$) & Old ($\uparrow$) & New ($\uparrow$) & Avg ($\uparrow$) & $A_{HM}$ ($\uparrow$) \\
\hline
\hline
\cellcolor{orange!15}Joint (Oracle) & 88.8 & 91.2 $\pm$ \scriptsize 0.4 & 37.5 $\pm$ \scriptsize 3.0 & 79.4 $\pm$ \scriptsize 0.7 & 91.4 $\pm$ \scriptsize 1.1 & 52.2 $\pm$ \scriptsize 4.2 & 77.3 $\pm$ \scriptsize 2.1 &  &  & \cellcolor{gray!20}\textbf{70.9 $\pm$ \scriptsize 1.2} & \textbf{62.4 $\pm$ \scriptsize 0.4} \\
\hline
FT & 88.8 & \phantom{0}0.0 $\pm$ \scriptsize 0.0 & 91.6 $\pm$ \scriptsize 3.7 & 20.3 $\pm$ \scriptsize 0.8 & \phantom{0}0.0 $\pm$ \scriptsize 0.0 & 69.5 $\pm$ \scriptsize 11.9 & 12.4 $\pm$ \scriptsize 2.1 & \cellcolor{white!20}\phantom{0}0.0 $\pm$ \scriptsize 0.0 & \cellcolor{white!30}85.8 $\pm$ \scriptsize 9.4 & 13.4 $\pm$ \scriptsize 1.5 & \phantom{0}0.0 $\pm$ \scriptsize 0.0 \\
FE & 88.8 & 56.2 $\pm$ \scriptsize 3.4 & 85.7 $\pm$ \scriptsize 6.5 & 62.7 $\pm$ \scriptsize 2.4 & 37.2 $\pm$ \scriptsize 6.7 & 64.3 $\pm$ \scriptsize 11.9 & 41.9 $\pm$ \scriptsize 6.9 & \cellcolor{white!20}17.5 $\pm$ \scriptsize 5.1 & \cellcolor{white!30}77.3 $\pm$ \scriptsize 8.8 & 26.8 $\pm$ \scriptsize 3.4 & 28.5 $\pm$ \scriptsize 6.4 \\
\cellcolor{gray!15}FE, Frozen & \cellcolor{gray!15}88.8 & 69.6 $\pm$ \scriptsize 4.1 & 77.4 $\pm$ \scriptsize 9.2 & \cellcolor{gray!15}71.3 $\pm$ \scriptsize 1.9 & \cellcolor{gray!15}61.9 $\pm$ \scriptsize 2.1 & 59.1 $\pm$ \scriptsize 11.7 & \cellcolor{gray!15}61.4 $\pm$ \scriptsize 2.7 & \cellcolor{gray!15}44.7 $\pm$ \scriptsize 3.2 & \cellcolor{gray!15}54.5 $\pm$ \scriptsize 6.7 & \cellcolor{gray!15}46.2 $\pm$ \scriptsize 2.7 & \cellcolor{gray!15}49.1 $\pm$ \scriptsize 4.3 \\
\hline
LWF \cite{li2017learning} & 88.8 & \phantom{0}0.0 $\pm$ \scriptsize 0.0 & 91.2 $\pm$ \scriptsize 6.2 & 20.2 $\pm$ \scriptsize 1.4 & \phantom{0}0.0 $\pm$ \scriptsize 0.0 & 70.3 $\pm$ \scriptsize 5.8 & 12.5 $\pm$ \scriptsize 1.0 & \phantom{0}0.0 $\pm$ \scriptsize 0.0 & \textbf{88.4 $\pm$ \scriptsize 13.7} & 13.8 $\pm$ \scriptsize 2.1 & \phantom{0}0.0 $\pm$ \scriptsize 0.0 \\
\hline
L2P \cite{wang2022learning}** & 88.8 & 17.3 $\pm$ \scriptsize 5.2 & 30.9 $\pm$ \scriptsize 25.9 & 20.3 $\pm$ \scriptsize 5.9 & 12.3 $\pm$ \scriptsize 5.9 & \phantom{0}2.1 $\pm$ \scriptsize 3.8 & 10.5 $\pm$ \scriptsize 4.8 & \phantom{0}8.2 $\pm$ \scriptsize 4.0 & \phantom{0}6.9 $\pm$ \scriptsize 8.5 & \phantom{0}7.9 $\pm$ \scriptsize 3.9 & \phantom{0}7.5 $\pm$ \scriptsize 5.5 \\
CODA-P \cite{smith2023coda}** & 87.7 & 19.9 $\pm$ \scriptsize 5.6 & \phantom{0}0.5 $\pm$ \scriptsize 0.7 & 15.6 $\pm$ \scriptsize 4.5 & 13.6 $\pm$ \scriptsize 4.0 & \phantom{0}2.2 $\pm$ \scriptsize 1.9 & 11.6 $\pm$ \scriptsize 1.9 & \phantom{0}7.9 $\pm$ \scriptsize 1.8 & 14.1 $\pm$ \scriptsize 21.4 & \phantom{0}8.8 $\pm$ \scriptsize 2.4 & 10.1 $\pm$ \scriptsize 3.2 \\
\hline
\cellcolor{gray!15}ALICE \cite{peng2022few} & \cellcolor{gray!15}92.1 & 86.0 $\pm$ \scriptsize 3.5 & 24.5 $\pm$ \scriptsize 14.9 & \cellcolor{gray!15}72.4 $\pm$ \scriptsize 5.7 & 72.1 $\pm$ \scriptsize 5.7 & 22.5 $\pm$ \scriptsize 16.9 & \cellcolor{gray!15}63.3 $\pm$ \scriptsize 7.6 & \cellcolor{gray!15}\textbf{62.5 $\pm$ \scriptsize 6.8} & \cellcolor{gray!15}11.9 $\pm$ \scriptsize 9.9 & \cellcolor{gray!15}\textbf{54.6 $\pm$ \scriptsize 6.9} & \cellcolor{gray!15}20.0 $\pm$ \scriptsize 8.1 \\
\hline
\cellcolor{gray!20}POET (Ours) & \cellcolor{gray!20}91.9  & 71.1 $\pm$ \scriptsize 4.7 & 80.8 $\pm$ \scriptsize 6.7 & \cellcolor{gray!20}73.2 $\pm$ \scriptsize 3.7 & 63.1 $\pm$ \scriptsize 2.2 & 56.7 $\pm$ \scriptsize 9.4 & \cellcolor{gray!20}61.9 $\pm$ \scriptsize 1.8 & \cellcolor{gray!20}45.9 $\pm$ \scriptsize 2.6 & \cellcolor{gray!20}72.4 $\pm$ \scriptsize 7.1 & \cellcolor{gray!20}50.0 $\pm$ \scriptsize 1.6 & \cellcolor{gray!20}\textbf{56.2 $\pm$ \scriptsize 1.6} \\
\bottomrule
\end{tabular}
\end{adjustbox}
\vspace{-10pt}
\end{table*}

\vspace{-12pt}
\section{Ablation Studies and Analysis}
\label{sec: ablations}
\vspace{-6pt}


\noindent \textbf{Importance of prompts in POET.} First, we \textit{consider the contribution of prompt offsets in POET}. Since we only attach prompts to address continual learning in POET, removing prompts gives the Feature Extraction (FE) baseline (`w/o prompts', Table \ref{tab: main ablation}) where the backbone is frozen after base training and only the classifier is expanded and updated on classification loss of new classes. POET improves both, `Old' ($\uparrow 20.1\%$) and `New' ($\uparrow 10.6\%$) marked \textcolor{blue}{in blue}. 



\begin{wraptable}[10]{r}{0.55\textwidth}
\vspace{-36pt}
    \centering
        \caption{\scriptsize \textbf{Prompt Selection Mechanism Analysis on NTU RGB+D dataset (\%, $\uparrow$):} `w/o' denotes removing that component from POET, numbers in brackets are wrt \textit{POET ($M=T$)} experiment. `Avg' accuracy is biased towards `Old' classes accuracy, $A_{HM}$ is good indicator of trade-off between `New' and `Old'.}
    \scalebox{0.53}{\begin{tabular}{@{}l||c|HHc|HHc|HHc|llll@{}}\toprule 
NTU RGB+D
& \normalsize $\boldsymbol{\mathcal{UB}^{(0)}}$
& \multicolumn{3}{c}{\normalsize \tabletask{1}} 
& \multicolumn{3}{c}{\normalsize \tabletask{2}}
& \multicolumn{3}{c}{\normalsize \tabletask{3}}
& \multicolumn{4}{c}{\normalsize \tabletask{4}} \\
\cmidrule(lr){2-2} \cmidrule(lr){3-5} \cmidrule(lr){6-8} \cmidrule(lr){9-11}  \cmidrule(l){12-15} 
\textbf{Method} & Base  & Old & New & Avg & Old & New & Avg & Old & New & Avg & Old & New & Avg & $A_{HM}$  \\
\hline
w/o prompts & 88.4 & 76.5 & 59.1 & 74.5 & 67.9 & 51.4 & 66.3 & 50.4 & 40.8 & 49.5 & \cellcolor{blue!10}39.2 \textcolor{red}{(-20.1)} & \cellcolor{blue!10}46.8 \textcolor{red}{(-10.6)} & \cellcolor{blue!10}39.9 & \cellcolor{blue!10}42.7 \\
w/o coupled optim. & 88.0 & 85.3 & 61.6 & 82.8 & 78.0 & 50.6 & 75.3 & 67.1 & 55.1 & 65.8 & 56.5 \textcolor{red}{(\phantom{0}-2.8)} & 51.3 \textcolor{red}{(\phantom{0}-6.1)} & 56.1 & 53.8 \\
w/o clustering loss & 85.5 & 86.6 & 41.5 & 81.6 & 78.9 & 36.8 & 74.3 & 69.2 & 28.9 & 64.5 & \textbf{62.0} \textcolor{blue}{(+2.7)} & 18.2 \textcolor{red}{(-39.2)} & 57.0 & 28.1 \\
w/o QA update & 87.9 & 84.7 & 66.3 & 82.8 & 80.9 & 45.5 & \textbf{77.4} & 70.1 & 59.5 & \textbf{69.1} & 59.4 \textcolor{blue}{(+0.1)} & 52.8 \textcolor{red}{(\phantom{0}-4.6)} & 58.7 & 55.9 \\
w/o sorting & 88.2 & 84.5 & 63.6 & 82.2 & 78.5 & 45.2 & 75.2 & 70.0 & 56.2 & 68.8 & 59.9 \textcolor{blue}{(+0.6)} & \cellcolor{blue!20}46.6 \textcolor{red}{(-10.8)} & 58.8 & 52.4 \\
\hline
POET ($M>T$) & 87.9 & 84.7 & 66.4 & 82.7 & 80.8 & 45.2 & 77.2 & 69.8 & 58.9 & 68.8 & 60.3 \textcolor{blue}{(+1.0)} & 54.4 \textcolor{red}{(\phantom{0}-3.0)} & \textbf{59.8} & 57.2 \\
\cellcolor{gray!20}POET ($M=T$) & \cellcolor{gray!20}87.9 & 84.9 & 65.6 & \cellcolor{gray!20}\textbf{82.8} & 80.3 & 45.8 & \cellcolor{gray!20}76.8 & 69.5 & 60.0 & \cellcolor{gray!20}68.6 & \cellcolor{gray!20}59.3 & \cellcolor{gray!20}\textbf{57.4} & \cellcolor{gray!20}\textbf{59.2} & \cellcolor{gray!20}\textbf{58.3} \\
\bottomrule
\end{tabular}
    }
    \label{tab: main ablation}
\end{wraptable}

\noindent \textbf{Prompt Selection Mechanism.} In Table \ref{tab: main ablation}, we investigate our prompt selection mechanism and optimization choices. The \textbf{`w/o coupled optim.'} experiment is a direct comparison of our additive \textit{prompt attachment} with the de-coupled optimization in L2P \cite{wang2022learning}. Updating key parameters but keeping only query adaptor $QA$ frozen after $\mathcal{UB}^{(0)}$ training (\textbf{`w/o QA update'}) reduces `New' only performance of \task{4} by $4.6\%$ as the query function stays fixed at base session learning and is not discriminative towards new classes. \textbf{`W/o clustering loss'} from Eq.~\ref{eqn: 6}, performance drops starting from $\mathcal{UB}^{(0)}$ itself. The only difference between the experiment \textbf{`w/o sorting'} and `POET (M=T)' is that we do not \textit{sort} the cosine similarity before selecting top $T$ indices (same as Fig \ref{fig: ordering_qual_vis}). The $10.8\% \uparrow$ in `New' performance validates that our prompt selection mechanism is learning to chose a distinct temporal ordering for prompt tuning of new input samples. With pool expansion (\textbf{`POET, $M>T$'}), we get more flexibility in the stability-plasticity trade-offs depending on how many new prompts we attach. For $R=6$, `Old' is improved. 
In Table \ref{tab: main ablation}, we keep POET's additive prompt attachment and only vary prompt selection.

\noindent \textbf{Prompt Attachment Mechanism.}
In Table \ref{tab: prompt attachment}, we keep our end-to-end optimization and ordered prompt selection as a constant and ablate prompt shape and attachment operator $f_p(.)$. Drawing a parallel with transformers which concatenate prompts along the token dimension, we conduct experiments concatenating prompts along the (i) temporal dimension of the skeleton input feature embedding $\mathbf{X_e}$ (\textit{`CONCAT temporal'}) and (ii) feature dimension $C_e$ (\textit{`CONCAT feature'}). We find that addition works better than concatenation and 
\begin{wraptable}[9]{r}{0.5\textwidth}
\vspace{-32pt}
    \centering
        \caption{\scriptsize \textbf{Prompt Attachment Analysis (\%, $\uparrow$):} The best prompt attachment choice $f_p(.)$ is \textit{Adding} \#prompts same as \#input frames (T=64).}
    \scalebox{0.5}{\begin{tabular}{@{}l||c|HHc|HHc|HHc|llll@{}} \toprule 
NTU RGB+D
& \normalsize $\boldsymbol{\mathcal{UB}^{(0)}}$
& \multicolumn{3}{c}{\normalsize \tabletask{1}} 
& \multicolumn{3}{c}{\normalsize \tabletask{2}}
& \multicolumn{3}{c}{\normalsize \tabletask{3}}
& \multicolumn{4}{c}{\normalsize \tabletask{4}} \\
\cmidrule(lr){2-2} \cmidrule(lr){3-5} \cmidrule(lr){6-8} \cmidrule(lr){9-11}  \cmidrule(l){12-15} 
\textbf{Method} & Base & Old & New & Avg & Old & New & Avg & Old & New & Avg & Old & New & Avg & $A_{HM}$  \\
\hline
CONCAT temporal, $T'=64$  & 88.6 & 69.7 & 75.6 & 70.3 & 62.5 & 60.2 & 62.4 & 48.7 & 60.0 & 49.8 & 33.6 & 50.5 & 35.1 & 40.3 \\
CONCAT feature, $T'=64$  & 87.7 & 85.4 & 58.4 & 82.4 & 79.2 & 41.8 & 75.5 & 68.2 & 54.2 & 66.9 & 57.1 & 41.5 & 56.0 & 48.1 \\
Cross Attention, $T'=64$ & 82.9 &  &  & 77.4 &  &  & 72.2 &  &  & 65.0 & 57.1 & 32.3 & 55.0 & 41.2 \\ 
\hline
ADD, $T'=1$ & \textbf{88.7} & 73.4 & 72.8 & 73.3 & 62.0 & 69.3 & 62.7 & 45.0 & 50.6 & 45.5 & 33.7 & 47.0 & 34.8 & 39.3 \\
\hline
\cellcolor{gray!20}ADD, $T'=64$ (Ours) & \cellcolor{gray!20}87.9 & 84.9 & \cellcolor{gray!20}65.6 & \cellcolor{gray!20}\textbf{82.8} & 80.3 & \cellcolor{gray!20}45.8 & \cellcolor{gray!20}\textbf{76.8} & 69.5 & \cellcolor{gray!20}60.0 & \cellcolor{gray!20}\textbf{68.6} & \cellcolor{gray!20}{59.3} & \cellcolor{gray!20}\textbf{57.4} & \cellcolor{gray!20}{59.2} & \cellcolor{gray!20}\textbf{58.3} \\
\bottomrule
\end{tabular}
    }
    \label{tab: prompt attachment}
\end{wraptable}

\vspace{-0.15in}
\noindent cross attention. We also verify our hypothesis that selecting the same number of prompts as the input temporal dimension ($T=64$ for NTU RGB+D and $T=8$ for SHREC-2017) yields better results as compared to adding the same prompt frame to each input embedding frame (\textit{`Addition $T'=1$'}).

\vspace{-7pt}
\section{Conclusions and Future Work}
\vspace{-6pt}
The problem of continually adapting human action models to new user categories over time has gained prominence with the rising availability of XR devices. However, this setting poses unique challenges: (i) the user may be able to provide only a few samples for training, and (ii) accessing data from earlier sessions may violate privacy considerations. We hence propose a method based on prompt offset tuning to address this problem in this work. Prompt tuning to address learning over newer tasks has been attempted in recent years. However, these works have: (1) typically been designed for image-based tasks, (2) relied on strongly pretrained transformer backbones, (3) required full supervision for new tasks, and (4) exclusively applied prompt tuning to transformer architectures. This work departs from these four characteristics. Our work demonstrates that prompt offset tuning is a promising option to evolve and adapt skeleton-based human action models to new user classes. The careful design of each component of the proposed methodology finds validation in the promising results across well-known skeleton-based action recognition benchmarks. Our ablation studies and analysis corroborate our design choices in our implementation. 
Looking ahead, it will be interesting to explore how our approach and its design choices adapt when a ``generalist backbone" trained on a large corpus of action recognition data becomes accessible. Extending our method for differential privacy is another interesting direction of future work.




%
%


\bibliographystyle{unsrt}  
\bibliography{main}

\begin{thebibliography}{10}

\bibitem{hinojosa2022privhar}
Carlos Hinojosa, Miguel Marquez, Henry Arguello, Ehsan Adeli, Li~Fei-Fei, and Juan~Carlos Niebles.
\newblock Privhar: Recognizing human actions from privacy-preserving lens.
\newblock In {\em European Conference on Computer Vision}, pages 314--332. Springer, 2022.

\bibitem{albrecht2016gdpr}
Jan~Philipp Albrecht.
\newblock How the gdpr will change the world.
\newblock {\em Eur. Data Prot. L. Rev.}, 2:287, 2016.

\bibitem{wang2022learning}
Zifeng Wang, Zizhao Zhang, Chen-Yu Lee, Han Zhang, Ruoxi Sun, Xiaoqi Ren, Guolong Su, Vincent Perot, Jennifer Dy, and Tomas Pfister.
\newblock Learning to prompt for continual learning.
\newblock In {\em Proceedings of the IEEE/CVF Conference on Computer Vision and Pattern Recognition}, pages 139--149, 2022.

\bibitem{ma2022learning}
Ning Ma, Hongyi Zhang, Xuhui Li, Sheng Zhou, Zhen Zhang, Jun Wen, Haifeng Li, Jingjun Gu, and Jiajun Bu.
\newblock Learning spatial-preserved skeleton representations for few-shot action recognition.
\newblock In {\em European Conference on Computer Vision}, pages 174--191. Springer, 2022.

\bibitem{zhu2023adaptive}
Anqi Zhu, Qiuhong Ke, Mingming Gong, and James Bailey.
\newblock Adaptive local-component-aware graph convolutional network for one-shot skeleton-based action recognition.
\newblock In {\em Proceedings of the IEEE/CVF Winter Conference on Applications of Computer Vision}, pages 6038--6047, 2023.

\bibitem{yue2022action}
Rujing Yue, Zhiqiang Tian, and Shaoyi Du.
\newblock Action recognition based on rgb and skeleton data sets: A survey.
\newblock {\em Neurocomputing}, 2022.

\bibitem{ren2020survey}
Bin Ren, Mengyuan Liu, Runwei Ding, and Hong Liu.
\newblock A survey on 3d skeleton-based action recognition using learning method.
\newblock {\em Cyborg and Bionic Systems}, 2020.

\bibitem{zhang2019comprehensive}
Hong-Bo Zhang, Yi-Xiang Zhang, Bineng Zhong, Qing Lei, Lijie Yang, Ji-Xiang Du, and Duan-Sheng Chen.
\newblock A comprehensive survey of vision-based human action recognition methods.
\newblock {\em Sensors}, 19(5):1005, 2019.

\bibitem{li2021else}
Tianjiao Li, Qiuhong Ke, Hossein Rahmani, Rui~En Ho, Henghui Ding, and Jun Liu.
\newblock Else-net: Elastic semantic network for continual action recognition from skeleton data.
\newblock In {\em Proceedings of the IEEE/CVF International Conference on Computer Vision}, pages 13434--13443, 2021.

\bibitem{Aich2023Data}
Shubhra Aich, Jesus Ruiz-Santaquiteria, Zhenyu Lu, Prachi Garg, K~J Joseph, Alvaro Fernandez, Vineeth~N Balasubramanian, Kenrick Kin, Chengde Wan, Nicati~Cihan Camgoz, Shugao Ma, and Fernando De~la Torre.
\newblock Data-free class-incremental hand gesture recognition.
\newblock In {\em Proceedings of the IEEE/CVF International Conference on Computer Vision}, 2023.

\bibitem{wang2023molo}
Xiang Wang, Shiwei Zhang, Zhiwu Qing, Changxin Gao, Yingya Zhang, Deli Zhao, and Nong Sang.
\newblock Molo: Motion-augmented long-short contrastive learning for few-shot action recognition.
\newblock In {\em Proceedings of the IEEE/CVF Conference on Computer Vision and Pattern Recognition}, pages 18011--18021, 2023.

\bibitem{lester2021power}
Brian Lester, Rami Al-Rfou, and Noah Constant.
\newblock The power of scale for parameter-efficient prompt tuning.
\newblock {\em arXiv preprint arXiv:2104.08691}, 2021.

\bibitem{li2021prefix}
Xiang~Lisa Li and Percy Liang.
\newblock Prefix-tuning: Optimizing continuous prompts for generation.
\newblock {\em arXiv preprint arXiv:2101.00190}, 2021.

\bibitem{jia2022visual}
Menglin Jia, Luming Tang, Bor-Chun Chen, Claire Cardie, Serge Belongie, Bharath Hariharan, and Ser-Nam Lim.
\newblock Visual prompt tuning.
\newblock In {\em European Conference on Computer Vision}, pages 709--727. Springer, 2022.

\bibitem{zhu2023prompt}
Beier Zhu, Yulei Niu, Yucheng Han, Yue Wu, and Hanwang Zhang.
\newblock Prompt-aligned gradient for prompt tuning.
\newblock In {\em Proceedings of the IEEE/CVF International Conference on Computer Vision}, pages 15659--15669, 2023.

\bibitem{razdaibiedina2023progressive}
Anastasia Razdaibiedina, Yuning Mao, Rui Hou, Madian Khabsa, Mike Lewis, and Amjad Almahairi.
\newblock Progressive prompts: Continual learning for language models.
\newblock {\em arXiv preprint arXiv:2301.12314}, 2023.

\bibitem{wang2022s}
Yabin Wang, Zhiwu Huang, and Xiaopeng Hong.
\newblock S-prompts learning with pre-trained transformers: An occam’s razor for domain incremental learning.
\newblock {\em Advances in Neural Information Processing Systems}, 35:5682--5695, 2022.

\bibitem{bowman2023carte}
Benjamin Bowman, Alessandro Achille, Luca Zancato, Matthew Trager, Pramuditha Perera, Giovanni Paolini, and Stefano Soatto.
\newblock a-la-carte prompt tuning (apt): Combining distinct data via composable prompting.
\newblock In {\em Proceedings of the IEEE/CVF Conference on Computer Vision and Pattern Recognition}, pages 14984--14993, 2023.

\bibitem{wang2022dualprompt}
Zifeng Wang, Zizhao Zhang, Sayna Ebrahimi, Ruoxi Sun, Han Zhang, Chen-Yu Lee, Xiaoqi Ren, Guolong Su, Vincent Perot, Jennifer Dy, et~al.
\newblock Dualprompt: Complementary prompting for rehearsal-free continual learning.
\newblock In {\em European Conference on Computer Vision}, pages 631--648. Springer, 2022.

\bibitem{smith2023coda}
James~Seale Smith, Leonid Karlinsky, Vyshnavi Gutta, Paola Cascante-Bonilla, Donghyun Kim, Assaf Arbelle, Rameswar Panda, Rogerio Feris, and Zsolt Kira.
\newblock Coda-prompt: Continual decomposed attention-based prompting for rehearsal-free continual learning.
\newblock In {\em Proceedings of the IEEE/CVF Conference on Computer Vision and Pattern Recognition}, pages 11909--11919, 2023.

\bibitem{tang2023prompt}
Yu-Ming Tang, Yi-Xing Peng, and Wei-Shi Zheng.
\newblock When prompt-based incremental learning does not meet strong pretraining.
\newblock {\em arXiv preprint arXiv:2308.10445}, 2023.

\bibitem{tao2020few}
Xiaoyu Tao, Xiaopeng Hong, Xinyuan Chang, Songlin Dong, Xing Wei, and Yihong Gong.
\newblock Few-shot class-incremental learning.
\newblock In {\em Proceedings of the IEEE/CVF Conference on Computer Vision and Pattern Recognition}, pages 12183--12192, 2020.

\bibitem{dong2021few}
Songlin Dong, Xiaopeng Hong, Xiaoyu Tao, Xinyuan Chang, Xing Wei, and Yihong Gong.
\newblock Few-shot class-incremental learning via relation knowledge distillation.
\newblock In {\em Proceedings of the AAAI Conference on Artificial Intelligence}, volume~35, pages 1255--1263, 2021.

\bibitem{peng2022few}
Can Peng, Kun Zhao, Tianren Wang, Meng Li, and Brian~C Lovell.
\newblock Few-shot class-incremental learning from an open-set perspective.
\newblock In {\em European Conference on Computer Vision}, pages 382--397. Springer, 2022.

\bibitem{zhou2022forward}
Da-Wei Zhou, Fu-Yun Wang, Han-Jia Ye, Liang Ma, Shiliang Pu, and De-Chuan Zhan.
\newblock Forward compatible few-shot class-incremental learning.
\newblock In {\em Proceedings of the IEEE/CVF conference on computer vision and pattern recognition}, pages 9046--9056, 2022.

\bibitem{hersche2022constrained}
Michael Hersche, Geethan Karunaratne, Giovanni Cherubini, Luca Benini, Abu Sebastian, and Abbas Rahimi.
\newblock Constrained few-shot class-incremental learning.
\newblock In {\em Proceedings of the IEEE/CVF Conference on Computer Vision and Pattern Recognition}, pages 9057--9067, 2022.

\bibitem{pernici2021class}
Federico Pernici, Matteo Bruni, Claudio Baecchi, Francesco Turchini, and Alberto Del~Bimbo.
\newblock Class-incremental learning with pre-allocated fixed classifiers.
\newblock In {\em 2020 25th International Conference on Pattern Recognition (ICPR)}, pages 6259--6266. IEEE, 2021.

\bibitem{yang2023neural}
Yibo Yang, Haobo Yuan, Xiangtai Li, Zhouchen Lin, Philip Torr, and Dacheng Tao.
\newblock Neural collapse inspired feature-classifier alignment for few-shot class-incremental learning.
\newblock In {\em The Eleventh International Conference on Learning Representations}, 2023.

\bibitem{dwivedi2022graph}
Vijay~Prakash Dwivedi, Anh~Tuan Luu, Thomas Laurent, Yoshua Bengio, and Xavier Bresson.
\newblock Graph neural networks with learnable structural and positional representations.
\newblock In {\em International Conference on Learning Representations}, 2022.

\bibitem{dwivedi2020generalization}
Vijay~Prakash Dwivedi and Xavier Bresson.
\newblock A generalization of transformer networks to graphs.
\newblock {\em arXiv preprint arXiv:2012.09699}, 2020.

\bibitem{mialon2021graphit}
Gr{\'e}goire Mialon, Dexiong Chen, Margot Selosse, and Julien Mairal.
\newblock Graphit: Encoding graph structure in transformers.
\newblock {\em arXiv preprint arXiv:2106.05667}, 2021.

\bibitem{chen2019construct}
Yuxiao Chen, Long Zhao, Xi~Peng, Jianbo Yuan, and Dimitris~N Metaxas.
\newblock Construct dynamic graphs for hand gesture recognition via spatial-temporal attention.
\newblock In {\em Proceedings of the British Machine Vision Conference (BMVC)}, 2019.

\bibitem{van2017neural}
Aaron Van Den~Oord, Oriol Vinyals, et~al.
\newblock Neural discrete representation learning.
\newblock {\em Advances in neural information processing systems}, 30, 2017.

\bibitem{bengio2013estimating}
Yoshua Bengio, Nicholas L{\'e}onard, and Aaron Courville.
\newblock Estimating or propagating gradients through stochastic neurons for conditional computation.
\newblock {\em arXiv preprint arXiv:1308.3432}, 2013.

\bibitem{williams2020hierarchical}
Will Williams, Sam Ringer, Tom Ash, David MacLeod, Jamie Dougherty, and John Hughes.
\newblock Hierarchical quantized autoencoders.
\newblock {\em Advances in Neural Information Processing Systems}, 33:4524--4535, 2020.

\bibitem{dhariwal2020jukebox}
Prafulla Dhariwal, Heewoo Jun, Christine Payne, Jong~Wook Kim, Alec Radford, and Ilya Sutskever.
\newblock Jukebox: A generative model for music.
\newblock {\em arXiv preprint arXiv:2005.00341}, 2020.

\bibitem{zheng2023online}
Chuanxia Zheng and Andrea Vedaldi.
\newblock Online clustered codebook.
\newblock In {\em Proceedings of the IEEE/CVF International Conference on Computer Vision}, pages 22798--22807, 2023.

\bibitem{liu2020learning}
Xuanqing Liu, Hsiang-Fu Yu, Inderjit Dhillon, and Cho-Jui Hsieh.
\newblock Learning to encode position for transformer with continuous dynamical model.
\newblock In {\em International conference on machine learning}, pages 6327--6335. PMLR, 2020.

\bibitem{li2021learnable}
Yang Li, Si~Si, Gang Li, Cho-Jui Hsieh, and Samy Bengio.
\newblock Learnable fourier features for multi-dimensional spatial positional encoding.
\newblock {\em Advances in Neural Information Processing Systems}, 34:15816--15829, 2021.

\bibitem{park2019deepsdf}
Jeong~Joon Park, Peter Florence, Julian Straub, Richard Newcombe, and Steven Lovegrove.
\newblock Deepsdf: Learning continuous signed distance functions for shape representation.
\newblock In {\em Proceedings of the IEEE/CVF conference on computer vision and pattern recognition}, pages 165--174, 2019.

\bibitem{rebuffi2018efficient}
Sylvestre-Alvise Rebuffi, Hakan Bilen, and Andrea Vedaldi.
\newblock Efficient parametrization of multi-domain deep neural networks.
\newblock In {\em Proceedings of the IEEE conference on computer vision and pattern recognition}, pages 8119--8127, 2018.

\bibitem{mallya2018packnet}
Arun Mallya and Svetlana Lazebnik.
\newblock Packnet: Adding multiple tasks to a single network by iterative pruning.
\newblock In {\em Proceedings of the IEEE conference on Computer Vision and Pattern Recognition}, pages 7765--7773, 2018.

\bibitem{rusu2016progressive}
Andrei~A Rusu, Neil~C Rabinowitz, Guillaume Desjardins, Hubert Soyer, James Kirkpatrick, Koray Kavukcuoglu, Razvan Pascanu, and Raia Hadsell.
\newblock Progressive neural networks.
\newblock {\em arXiv preprint arXiv:1606.04671}, 2016.

\bibitem{shahroudy2016ntu}
Amir Shahroudy, Jun Liu, Tian-Tsong Ng, and Gang Wang.
\newblock Ntu rgb+ d: A large scale dataset for 3d human activity analysis.
\newblock In {\em Proceedings of the IEEE conference on computer vision and pattern recognition}, pages 1010--1019, 2016.

\bibitem{10.2312:3dor.20171049}
Quentin~De Smedt, Hazem Wannous, Jean-Philippe Vandeborre, J.~Guerry, B.~Le Saux, and D.~Filliat.
\newblock {3D Hand Gesture Recognition Using a Depth and Skeletal Dataset}.
\newblock In Ioannis Pratikakis, Florent Dupont, and Maks Ovsjanikov, editors, {\em Eurographics Workshop on 3D Object Retrieval}. The Eurographics Association, 2017.

\bibitem{chen2021channel}
Yuxin Chen, Ziqi Zhang, Chunfeng Yuan, Bing Li, Ying Deng, and Weiming Hu.
\newblock Channel-wise topology refinement graph convolution for skeleton-based action recognition.
\newblock In {\em Proceedings of the IEEE/CVF International Conference on Computer Vision}, pages 13359--13368, 2021.

\bibitem{villa2023pivot}
Andr{\'e}s Villa, Juan~Le{\'o}n Alc{\'a}zar, Motasem Alfarra, Kumail Alhamoud, Julio Hurtado, Fabian~Caba Heilbron, Alvaro Soto, and Bernard Ghanem.
\newblock Pivot: Prompting for video continual learning.
\newblock In {\em Proceedings of the IEEE/CVF Conference on Computer Vision and Pattern Recognition}, pages 24214--24223, 2023.

\bibitem{ridnik2021imagenet}
Tal Ridnik, Emanuel Ben-Baruch, Asaf Noy, and Lihi Zelnik-Manor.
\newblock Imagenet-21k pretraining for the masses.
\newblock {\em arXiv preprint arXiv:2104.10972}, 2021.

\bibitem{radford2021learning}
Alec Radford, Jong~Wook Kim, Chris Hallacy, Aditya Ramesh, Gabriel Goh, Sandhini Agarwal, Girish Sastry, Amanda Askell, Pamela Mishkin, Jack Clark, et~al.
\newblock Learning transferable visual models from natural language supervision.
\newblock In {\em International conference on machine learning}, pages 8748--8763. PMLR, 2021.

\bibitem{li2017learning}
Zhizhong Li and Derek Hoiem.
\newblock Learning without forgetting.
\newblock {\em IEEE transactions on pattern analysis and machine intelligence}, 40(12):2935--2947, 2017.

\bibitem{kirkpatrick2017overcoming}
James Kirkpatrick, Razvan Pascanu, Neil Rabinowitz, Joel Veness, Guillaume Desjardins, Andrei~A Rusu, Kieran Milan, John Quan, Tiago Ramalho, Agnieszka Grabska-Barwinska, et~al.
\newblock Overcoming catastrophic forgetting in neural networks.
\newblock {\em Proceedings of the national academy of sciences}, 114(13):3521--3526, 2017.

\bibitem{hou2019learning}
Saihui Hou, Xinyu Pan, Chen~Change Loy, Zilei Wang, and Dahua Lin.
\newblock Learning a unified classifier incrementally via rebalancing.
\newblock In {\em Proceedings of the IEEE/CVF conference on computer vision and pattern recognition}, pages 831--839, 2019.

\bibitem{chaudhry2018riemannian}
Arslan Chaudhry, Puneet~K Dokania, Thalaiyasingam Ajanthan, and Philip~HS Torr.
\newblock Riemannian walk for incremental learning: Understanding forgetting and intransigence.
\newblock In {\em Proceedings of the European conference on computer vision (ECCV)}, pages 532--547, 2018.

\bibitem{chaudhry2018efficient}
Arslan Chaudhry, Marc'Aurelio Ranzato, Marcus Rohrbach, and Mohamed Elhoseiny.
\newblock Efficient lifelong learning with a-gem.
\newblock {\em arXiv preprint arXiv:1812.00420}, 2018.

\bibitem{li2023stprivacy}
Ming Li, Xiangyu Xu, Hehe Fan, Pan Zhou, Jun Liu, Jia-Wei Liu, Jiahe Li, Jussi Keppo, Mike~Zheng Shou, and Shuicheng Yan.
\newblock Stprivacy: Spatio-temporal privacy-preserving action recognition.
\newblock In {\em Proceedings of the IEEE/CVF International Conference on Computer Vision}, pages 5106--5115, 2023.

\bibitem{kumawat2022privacy}
Sudhakar Kumawat and Hajime Nagahara.
\newblock Privacy-preserving action recognition via motion difference quantization.
\newblock In {\em European Conference on Computer Vision}, pages 518--534. Springer, 2022.

\bibitem{lacoste2019quantifying}
Alexandre Lacoste, Alexandra Luccioni, Victor Schmidt, and Thomas Dandres.
\newblock Quantifying the carbon emissions of machine learning.
\newblock {\em arXiv preprint arXiv:1910.09700}, 2019.

\end{thebibliography}

\newpage
\clearpage
\appendix
\setcounter{page}{1}
\title{Supplementary Materials}
\subtitle{POET: \underline{P}rompt \underline{O}ffs\underline{e}t \underline{T}uning for Continual \\ Human Action Adaptation}
\author{Prachi Garg\inst{1} \and
Joseph K J\inst{3} \and
Vineeth N Balasubramanian\inst{3} \and 
Necati Cihan Camgoz\inst{2} \and
Chengde Wan\inst{2} \and
Kenrick Kin\inst{2} \and
Weiguang Si\inst{2} \and
Shugao Ma\inst{2} \and
Fernando De La Torre\inst{1}
}

\institute{Carnegie Mellon University, USA \and
Meta Reality Labs \and
Indian Institute of Technology, Hyderabad}

\maketitle

In this supplementary material, we provide the following additional information which we could not include in the main paper due to space constraints. We provide all implementation details (besides the code) herein.

\section*{Table of Contents}
\begin{enumerate}[label=(\Alph*)]
    \item \textbf{Implementation Details} 
    \begin{enumerate}[label=(\arabic*)]
        \item Training Details 
        \item Prompt Instantiation in Base Session $t=0$ (Algorithm 2)
        \item Classifier Update Protocol, $t>0$ 
        \item Prompt Attachment Analysis 
        \item Additional Dataset Details
        \item Adaptation of Baseline Methods to Problem Setting
    \end{enumerate}
    \item \textbf{Additional Results} 
    \begin{enumerate}[label=(\arabic*)]
        \item POET's Effectiveness in Learning New Knowledge while Mitigating Catastrophic Forgetting
        \begin{enumerate}
        \item How does POET mitigate catastrophic forgetting?
            \item Backward Forgetting Metric (BWF)
            \item Effect of variation of number of few-shots for training
        \end{enumerate}
        \item Impact of Prompts in POET 
        \item Stability-Plasticity Performance Trade-offs 
        \item Robustness to continual class order in user sessions
        \item Ordered Key Index Selection $(s_i)_{i=1}^{T}$: Qualitative Analysis 
        \item Prompt Pool Expansion (Algorithm 3) 
    \end{enumerate}
    \item \textbf{Broader Impact and Limitations}
\end{enumerate}



\clearpage
\section{Implementation Details}

\subsection{Training Details}
We use the same hyperparameters across all experiments for the NTU RGB+D activity recognition benchmark (in Tables \ref{sota activity}, \ref{tab: main ablation}, \ref{tab: prompt attachment} of the main paper, and Supplementary Table \ref{tab: classifier analysis, NTU} and Figure \ref{fig: old-new-ntu}). In the SHREC 2017 gesture benchmark also, all our experiments follow the exact same hyperparameter combination and learning strategy in Table \ref{SOTA gesture} and Figure \ref{fig: old-new-shrec}.

\vspace{4pt}
\noindent \textit{Activity recognition on NTU RGB+D benchmark.} In the base session $\mathcal{UB}^{(0)}$, we train the CTR-GCN \cite{chen2021channel} backbone for 50 epochs with initial LR=0.1. We use a batch size of 64 as in the original paper. Every continual user session $\mathcal{US}^{(t)}$ is trained for 5 epochs with an initial LR=0.1. 

\vspace{4pt}
\noindent \textit{Gesture recognition on SHREC 2017.} We train DG-STA \cite{chen2019construct} model in the base session $\mathcal{UB}^{(0)}$ for 300 epochs and initial LR=0.001, using batch size 32 and dropout set to 0.2 (default hyperparameters from the DG-STA paper). It is updated for 30 epochs in each user session $\mathcal{US}^{(t)}$, starting with initial LR=0.01. 

\vspace{4pt}
\par We select higher initial LRs in continual sessions $t>0$ because starting with a lower learning rate as compared to base session (as is standard practice in continual learning to prevent catastrophic forgetting) renders limited plasticity and the model is completely unable to learn new knowledge. Our choice enables learning of new knowledge from the few user samples, and we can study the model's stability-plasticity trade-offs, optimizing for a balance between the two. The continual session learning rates given above are used to update the (i) the classifier \textcolor{RedOrange}{$f_c$}, (ii) selected prompts in \textcolor{RedOrange}{$\mathbf{P}$}, and (iii) selected keys in \textcolor{RedOrange}{$\boldsymbol{K}$}. But for updating query adaptor \textcolor{RedOrange}{$f_{QA}$}, we use a learning rate of 0.01, in $t>0$ for both benchmarks. At the same time, we freeze all other layers in the query model. We find this adapts query adaptor to new tasks without overwriting existing base knowledge. In each continual session $t>0$, we use a batch size of 25 for NTU RGB+D, as there are 5 new classes each having 5 training samples (single batch per epoch). Similarly, batch size is 10 for SHREC 2017 with 2 new classes with 5 training samples each. For the clustering loss coefficient in Eq. \ref{eqn: 6}, we use $\lambda = 0.1$ for all experiments. 

\vspace{-0.4cm}
\begin{algorithm*}[bh!]
\vspace{-2pt}
\caption{\footnotesize Initialization \& Training of Prompts, $t = 0$}\label{alg:POET t=0}
\begin{algorithmic}
\State \textbf{Input:} Model $f^P(.) = f^P_c \circ f^P_g \circ f^P_e(.)$, \textit{pretrained} only on base $\mathcal{UB}^{(0)}$ data. 
\State \textbf{Initialize:} 
    \State \hskip0.8em 1. Main model $f$ as: $f_e \leftarrow f^P_e, f_g \leftarrow f^P_g, f_c \leftarrow f^P_c$.
    \State \hskip0.8em 2. Prompt pool $\mathbf{P}=\{\boldsymbol{P_j}\}_{j=1}^T$, Keys $\boldsymbol{K}=\{\boldsymbol{k_j}\}_{j=1}^T$ from $\mathcal{U}(0, 1)$.
    \State \hskip0.8em 2. Query function model $f_q$ as: $f_e' \leftarrow f^P_e, f_g' \leftarrow f^P_g$. $f_{QA}$ is randomly initialized.
    
\State \textbf{Freeze:} Query function layers \textcolor{cyan}{$f_g', f_e'$}.
\State \textbf{for} epochs and batch in base dataset ${(\mathbf{X}_i^0, \mathbf{y}_i^0)}_{i=1}^{|\mathcal{D}^{(0)}|}$ do
    \State \hskip0.8em 1. Get query feature $\boldsymbol{q}$ (Eq. \ref{eqn: 4}) ; Compute $\gamma(.)$ b/w query $\boldsymbol{q}$ and keys $\boldsymbol{K}$
    \State \hskip0.8em 2. \textcolor{black}{Sort} $\gamma(.)$; Get \textcolor{black}{ordered key index sequence} $(s_i)_{i=1}^{T}$ (Eq. \ref{eqn: 7})
    \State \hskip0.8em 3. Read pool memory $\mathbf{P}$ in order $(s_i)_{i=1}^{T}$ $\rightarrow$ Get prompt offsets $\mathbf{P_T}$
    \State \hskip0.8em 4. Get $\mathbf{X_e}$; \textbf{Add} $\mathbf{P_T}$ to it (Eq. \ref{eqn: 8}); get prediction $\mathbf{y}$ from prompted input (Eq. \ref{eqn: 1})
    \State \hskip0.8em 5. Use cross entropy loss (Equation \ref{eqn: 5}) to \textbf{update} prompt associated parameters \textcolor{RedOrange}{$f_{QA}, \boldsymbol{K}, \mathbf{P}$} and all main model parameters \textcolor{RedOrange}{$f_g, f_c, f_e$}.
    \State \hskip0.8em 6. Use clustering loss (Equation \ref{eqn: 6}) to \textbf{update} \textcolor{RedOrange}{$f_{QA}$, $\boldsymbol{K}$}. 
\State \textbf{end}
\State \textbf{Freeze:} Main model feature extractor $f_g$, input embedding layer $f_e$ for time $t>0$.
\vspace{-2pt}
\end{algorithmic}
\end{algorithm*}

\subsection{Base Session $\mathcal{UB}^{(0)}$: Prompt Instantiation and Training}
\label{subsection: prompt dimensions}
\noindent \textbf{Prompt instantiation, CTR-GCN:} CTR-GCN is a spatio-temporal graph convolutional network architecture with 10 multi-scale temporal convolutional (TCN-GCN) layers followed by an average pool over the spatial and temporal dimensions, and a final linear classification layer. The output feature dimensionality of the first four layers (L1-L4) is 64 channels, next four (L5-L8) is 128 and final two layers (L9 and L10) have 256 channels. An input skeleton sequence has $64$ temporal frames, each consisting of $25$ body joints, such that $x \in \mathcal{R}^{64 \times 25 \times 3}$. 

\par The input embedding after layer L1 is $x_e \in \mathcal{R}^{64 \times 25 \times 64}$, such that $C_e = 64$. We start from a prompt pool $\mathbf{P}=\{\boldsymbol{P_j}\}_{j=1}^T$ of size $M=T=64$. Each prompt in the pool $P_j \in \mathcal{R}^{25 \times 64}$ is designed to match the spatial and feature dimensions of the input embedding $x_e$. There are $M=64$ keys, each having feature dimension $\boldsymbol{k_j} \in \mathcal{R}^{64}$. Query adaptor $f_{QA}$ maps a feature embedding of size 256 (from the last layer of query model $f_g'$) to $C_e=64$ for feature dimension compatibility with the keys and prompts. We select $T=64$ prompts from the pool. After instantiating the prompt and key parameters, we train the prompt pool \textcolor{RedOrange}{$\mathbf{P}$}, keys \textcolor{RedOrange}{$\boldsymbol{K}$}, query adaptor \textcolor{RedOrange}{$f_{QA}$}, along with all main model parameters \textcolor{RedOrange}{$f_g, f_c, f_e$} on the base session data $\mathcal{D}^{(0)}$ as per Algorithm \ref{alg:POET t=0}.

\vspace{4pt}
\noindent \textbf{Prompt instantiation, DG-STA:} DG-STA is a fully connected graph transformer architecture with multi-head spatial and temporal attention layers. For every input skeleton hand gesture sequence, we use 8 temporal frames, each having 22 hand joint coordinates such that input is $x \in \mathcal{R}^{8 \times 22 \times 3}$. Output of the transformer's input embedding layer $f_e$ is $x_e \in \mathcal{R}^{8 \times 22 \times 128}$. As DG-STA expects a fully connected spatio-temporal graph across all joints in all frames, this is reshaped to a size $x_e \in \mathcal{R}^{176 \times 128}$ before passing it to the first attention layer of the transformer. We add our prompt to this reshaped embedding. We start from a pool of size $M=8$ prompts. As DG-STA is a transformer architecture, the output feature dimensionality remains constant (at 128) and the query adaptor input and output dimensions are the same ($C_e = 128$). The base session here is also trained using Algorithm \ref{alg:POET t=0}.

\subsection{Classifier Update Protocol in $\mathcal{US}^{(t)}, t>0$}
\label{setion: classifier update protocol}
Existing related work such as in few-shot class-incremental learning learn non-parametric classifiers by extracting class-mean prototypes, and do not expand the classifier with any new weights. However, we need to expand and train the classifier in order to obtain the error gradients for updating the prompts. We observe that a key source of forgetting is from the classifier as the logits tend to become biased towards the few-training samples of new classes. For SHREC/DG-STA, we observe that fine-tuning the entire classifier leads to a significant drop in performance of old classes (experiment `FE' in Table \ref{SOTA gesture} of main paper). This is because the frozen backbone is trained only on 1146 training samples from the 8 base session classes. Hence, the SHREC backbone exhibits low stability for retaining old knowledge, when updated on new data in subsequent sessions $t>0$. To alleviate this, we use a simple classifier update trick wherein after expanding the classifier in every continual session $t>0$, we turn the gradients of the old parameters in the classifier to zero before the error backpropagation. This trick has also been shown to work in prior continual prompt tuning works L2P \cite{wang2022learning}, CODA-P \cite{smith2023coda}. 

\par We observe that this trick proves sub-optimal in the NTU RGB+D (CTR-GCN) dataset. In Table \ref{tab: classifier analysis, NTU}, we evaluate performance on each usr session individually after learning every new user session. We observe that both, freezing (`\textbf{Regular, Freeze}') or finetuning (`\textbf{Regular, Tune}') of old class parameters in classifier suffers from poor stability-plasticity trade-offs. \textit{We observe that the intermediate tasks (\task{1}, \task{2}, \task{3}) learnt using few shot data particularly suffer severe forgetting as soon as the next task is learnt.} This is not the case for \task{0} performance in incremental sessions $t>0$ because the base feature extractor is trained on sufficient data in the activity benchmark (26,731 samples) and frozen henceforth, retaining performance on the base task \task{0} even in $t>0$. We call this the \textbf{`New-old forgetting'}. To address the biasing of logits towards new classes, we replace the main model's linear classification layer $f_c$ with a cosine normalization classifier $\theta_c^T$ as:
\begin{equation}
    p_{(x)} = \frac{exp(\eta <\theta_{c_i}^T f_g(x)>)}{\Sigma_j exp(\eta <\theta_{c_j}^T f_g(x)>)}
\end{equation}
where $\eta$ is a learnable scale parameter we learn in the base session and freeze in subsequent sessions. Also, in incremental sessions, we initialize the new-class parameters in the classifier as mean of previous class parameters. Notice the significant boost in performance across all tasks \task{i} in the cosine normalized classifier in Table \ref{tab: classifier analysis, NTU}.     

\begin{table*}[t!]
\setlength{\tabcolsep}{3pt}
\centering
\caption{\textbf{Experimenting with different classifiers for NTU RGB+D.} Here, we report the task-specific accuracy for each task the model has learnt so far, after learning every new task. Notice the sharp \textbf{forgetting of the \textit{intermediate} `New-Old'} tasks in regular classifier and our improvement using a cosine normalized classifier. Note, we are using a dynamically expanding parametric classifier in all experiments.}
\label{tab: classifier analysis, NTU}
\begin{adjustbox}{width=0.99\textwidth}
\begin{tabular}{lllllllllllllllll} \toprule 
& \task{0} & \multicolumn{2}{c}{\task{1}} & \multicolumn{3}{c}{\task{2}} & \multicolumn{4}{c}{\task{3}} & \multicolumn{5}{c}{\task{4}} \\
\cmidrule(lr){2-2} \cmidrule(lr){3-4} \cmidrule(lr){5-7} \cmidrule(lr){8-11}  \cmidrule(l){12-17} 
Activity & Base & \task{0} & \cellcolor{magenta!15}\task{1} & \task{0} & \cellcolor{magenta!15}\task{1} & \task{2} & \task{0} & \cellcolor{magenta!15}\task{1} & \task{2} & \cellcolor{blue!15}\task{3} & \task{0} & \cellcolor{magenta!15}\task{1} & \task{2} & \cellcolor{blue!15}\task{3} & \task{4} & \cellcolor{gray!15}\textbf{Avg} \\
\hline
Regular, Freeze & 89.2 & \cellcolor{white!20}73.5 & \cellcolor{magenta!15}\textbf{72.5} & \cellcolor{white!20}71.3 & \cellcolor{magenta!15}2.1 & \cellcolor{white!30}\textbf{63.8} & \cellcolor{white!20}65.3 & \cellcolor{magenta!15}0.0 & \cellcolor{white!20}0.0 & \cellcolor{blue!15}52.9 & \cellcolor{white!20}58.7 & \cellcolor{magenta!15}0.0 & \cellcolor{white!20}0.0 & \cellcolor{blue!15}0.0 & \cellcolor{white!30}48.9 & 43.1 \cellcolor{gray!15}\textcolor{red}{(-16.1)} \\
Regular, Tune & 89.2 & \cellcolor{white!20}80.9 & \cellcolor{magenta!15}64.9 & \cellcolor{white!20}67.4 & \cellcolor{magenta!15}22.1 & \cellcolor{white!30}34.4 & \cellcolor{white!20}57.2 & \cellcolor{magenta!15}6.32 & \cellcolor{white!20}24.3 & \cellcolor{blue!15}26.9 & \cellcolor{white!20}45.0 & \cellcolor{magenta!15}4.6 & \cellcolor{white!20}\textbf{20.9} & \cellcolor{blue!15}13.3 & \cellcolor{white!30}21.4 & 35.2 \cellcolor{gray!15}\textcolor{red}{(-24.0)} \\
\hline
Cosine (Ours) & 87.9 & \cellcolor{white!20}\textbf{84.9} & \cellcolor{magenta!15}65.3 & \cellcolor{white!20}\textbf{83.2} & \cellcolor{magenta!15}\textbf{56.0} & \cellcolor{white!30}45.8 & \cellcolor{white!20}\textbf{78.2} & \cellcolor{magenta!15}\textbf{36.3} & \cellcolor{white!20}\textbf{34.5} & \cellcolor{blue!15}\textbf{60.0} & \cellcolor{white!20}\textbf{71.3} & \cellcolor{magenta!15}\textbf{18.5} & \cellcolor{white!20}19.8 & \cellcolor{blue!15}\textbf{46.2} & \cellcolor{white!30}\textbf{57.4} & \cellcolor{gray!15}\textbf{59.2} \\
\bottomrule
\end{tabular}
\end{adjustbox}
\vspace{-10pt}
\end{table*}

\subsection{Analyzing Where to Attach Prompts}
\begin{figure*}[th]
\centering
\includegraphics[width=0.5\textwidth]{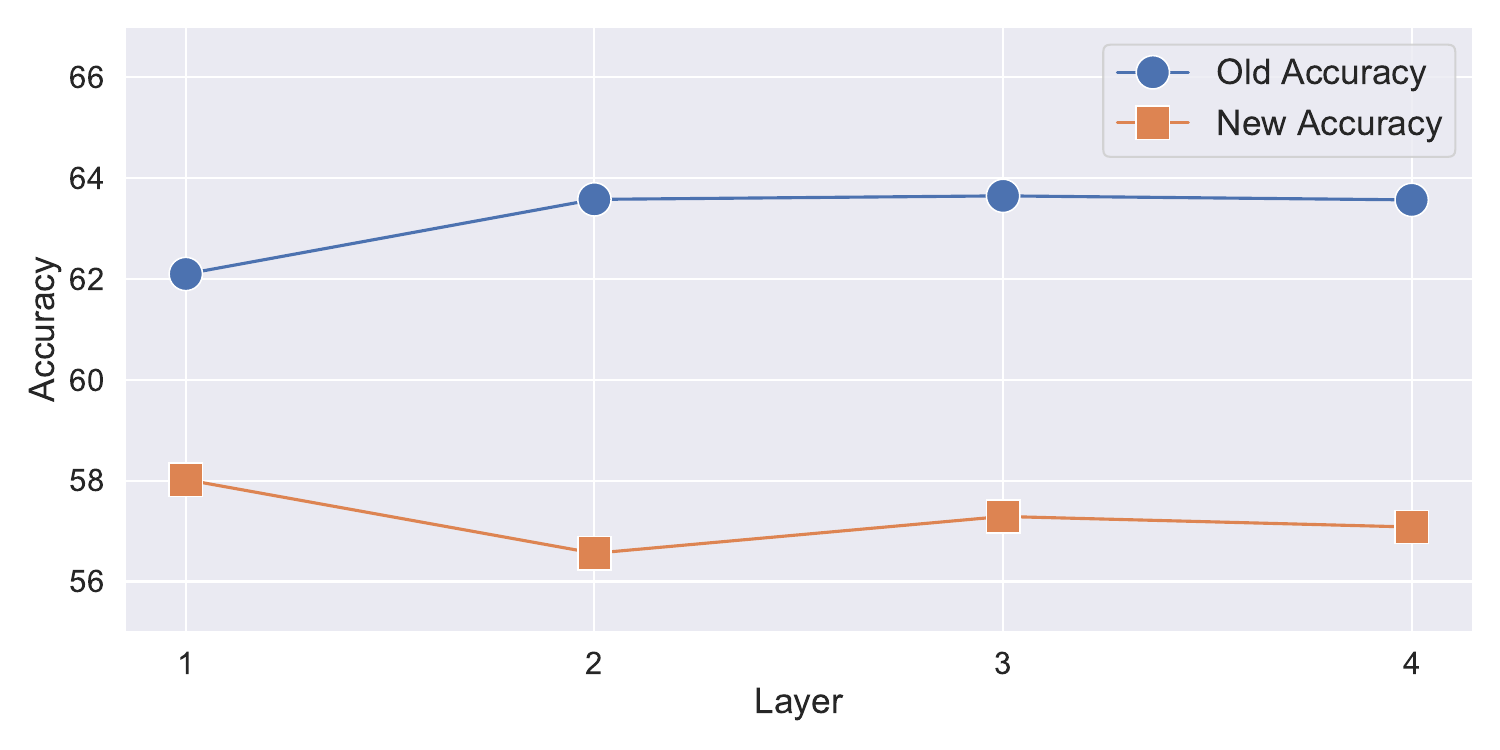}
\caption{Empirical analysis to study the impact of the layer at which our prompt is attached. Y-axis shows `Old' and `New' classes accuracy after Task 4 (after learning all 60 classes). We add a prompt of size $P_{T'} \in \mathcal{R}^{64, 25, 64}$ to different layers $\{1, 2, 3, 4\}$ of CTR-GCN, evaluated on the NTU RGB+D validation set. We select layer L1 due to its high performance on new classes.}
\label{fig: layer ablation}
\end{figure*}

We study the impact of attaching our prompt offsets at different main model layers in CTR-GCN in Figure \ref{fig: layer ablation}. As mentioned in Sec \ref{subsection: prompt dimensions}, output feature dimensionality (number of channels) of the first four layers in CTR-GCN is the same, $C_e=64$. We desire that the feature dimension size of the (i) prompt parameters, (ii) key parameters, and (iii) query adaptor be consistent with the main model feature embedding dimension at the layer being prompted. Hence, we only prompt the first four layers for a fair comparison as results may vary with variation in size of feature dimension being prompted. We created a separate $30\%$ validation set from the training data of $t \geq 1$ classes for this analysis. Our findings in Figure \ref{fig: layer ablation} indicate that the highest `New' task performance at the end of all four user sessions is achieved by prompting layer L1. We select layer L1 across all experiments in the paper.

\subsection{Additional Dataset Details}
The NTU RGB+D dataset has been collected from Microsoft Kinect V2 sensors from three different camera viewpoints by annotating 40 human users. The 60 classes consist of 40 daily action categories (drinking water, reading, writing, etc.), 9 health-related actions (coughing, sneezing, headache, etc.), and 11 mutual actions (handshaking, pushing another person, walking towards another person, etc.). NTU RGB+D has 40,320 training and 16,560 testing samples across 60 classes. We use the first 40 daily action classes as the base model data, and update on $5 \times 5 = 25$ training samples in each of the 4 incremental session. The base model is trained on 26731 training samples from the 40 base classes. 

The SHREC 2017 dataset has 14 fine-grained hand gestures captured using the short-range Intel Real Sense Depth, from 28 human subjects in second-person view. There are 1980 training and 840 testing samples. The base model for this benchmark is trained on 1146 training samples from 8 classes, and updated on $2 \times 5 = 10$ training samples in each of the 3 incremental session. 

\par Given the relatively small sizes of both datasets, we follow a class-incremental setting in our work, viz., our user sessions include new classes over time, not necessarily new users. 
Evaluating our work on user-specific continual streams is left as future work for larger datasets where this is feasible. 

\vspace{4pt}
\noindent \textbf{Class splits in user sessions.} There are two existing continual benchmarks for 3D skeleton-based action recognition: The experimental protocol of NTU RGB+D based class incremental benchmark \cite{li2021else} involves a single continual session, learning $50$ classes in the base session, and $10$ new classes in a single-incremental session. We consider this to be limiting and too simplistic to study our problem setting on. Moreover, their code is not publicly available. On the other hand, the more recent data-free class incremental learning for hand gesture benchmark \cite{Aich2023Data} learns 8 classes in base session and updates the model on a single class over 6 incremental sessions. We believe that this too does not lend itself to our setting, when there are only 5-shots for training each continual class and the base model is trained on small-scale data. 

\subsection{Adaptation of Baselines to Problem Setting}
\noindent \textbf{Adapting Learning to Prompt (L2P):}
We experiment with selecting $T'=4$ and $T'=64$ (same as POET) for this comparison on CTR-GCN backbone. We find the results with 4 prompts marginally better, hence we report those in Table 3 of main paper. We experiment with both temporal concatenation and spatio-temporal concatenation followed by remapping. For DG-STA, we select 8 prompts from a pool of size $M=10$, each prompt (22, 128); and concatenate this prompt of size (8, 22, 128) along the spatio-temporal dimension 176 of the input embedding (176, 128). We map this back to (176, 128) using a fully connected layer. In experiments where we remap using FC layer, we update this layer as well in future incremental sessions. To update the expanded classifier, we make logits of previous classes -inf, same as the classifier training protocol used in L2P, Dual-P and CODA-P.  

\vspace{4pt}
\noindent \textbf{Adapting CODA-P:} For activity recognition on CTR-GCN backbone, we construct a 4 dimensional CODA-prompt of size (100, T', 25, 64), such that the prompt component dimension of 100 gets collapsed after weighing and we can concatenate prompt of size (T', 25, 64) along the temporal or rolled out spatio-temporal dimension (same as L2P). The size of memory buffer (T') is kept consistent with L2P experiments. 

For gesture recognition on DG-STA backbone, \cite{smith2023coda} tunes a ViT-B/16 pre-trained on ImageNet-1K architecture instead of prompt tuning. This refers to concatenating half of the prompt to $K$ and $V$ of the MSA layer instead of concatenating along the token dimension. However, we don't have the luxury to modify the input embedding size or assume that the backbone is a transformer. Also for a fair comparison with L2P, we concatenate a fixed sized prompt to the input embedding and use a fully connected layer to map the feature dimension back to default input embedding size of (176, 128). Hence, sizes are as follows: initial prompt size (100, 5, 128), Attention (100, 128), Key (100, 128), each have 100 prompt components. After alpha weighting, fixed sized prompt P (5, 128) gets concatenated along joint dimension. We don't update query adaptor in this experiment. We implement all loss functions, including the orthogonality loss as is. Also, note that we attach prompts only at the input embedding layer for fair comparison of the prompting strategy. 

\noindent \textbf{Adapting ALICE \cite{peng2022few}:} For training the base session, we add a projection head to the feature extractor before the classification layer. Like the paper, we use two augmentations of every input and losses from the two augmentations are averaged before backpropagation. We use angular penalty for training the classifier. After base session learning, the projection head and classification layer are discarded, only learning feature extractor. Next, we use cosine distance and class-wise mean to generate class-prototypical feature vectors from the feature extractor's output. These prototypes are used for nearest mean classification. For incremental steps, no training is done. Only new class means are computed and evaluation is performed. 

\noindent \textbf{Adapting LUCIR \cite{hou2019learning} and EWC \cite{kirkpatrick2017overcoming}:}
We initialize model from previous checkpoint, such that classifier has random weights for new classes and previous classifier weights are copied over to previous class parameters in the classifier. Cross entropy loss is computed between logits from all the classes and current task ground truth labels. All regularization loss terms are implemented as proposed in their respective papers. For LwF \cite{li2017learning}, we use a $\lambda = 1.0$. 

\section{Additional Results}

\subsection{POET's Effectiveness in Learning New Knowledge while Mitigating Catastrophic Forgetting}

\vspace{-10pt}
\begin{table}[ht]
\centering
\caption{\textbf{POET Ablation Table:} We exhaustively study the contribution of various POET components towards mitigating forgetting of old knowledge (Old) as well as learning new knowledge (New).}
\resizebox{0.999\columnwidth}{!}{%
\setlength{\tabcolsep}{2pt}
\begin{tabular}{@{}p{5.2cm}|c|p{1.55cm}|c||c|llll@{}} \toprule
        & Prompt & Prompt Selection & Prompt Integration &
       \normalsize $\boldsymbol{\mathcal{UB}^{(0)}}$ & 
       \multicolumn{4}{c}{\normalsize \tabletask{4}} \\
       \cmidrule(lr){5-5} \cmidrule(lr){6-9} 
\textbf{Ablation} & &  & (Operator, \#Prompts) & Base ($\uparrow$)   & Old ($\uparrow$) & New ($\uparrow$) & Avg ($\uparrow$) & $A_{HM}$ ($\uparrow$) \\
\hline
\\[-1.8ex]
\cellcolor{gray!30}\textbf{POET (Ours)} &  \cellcolor{gray!30}\checkmark &  \cellcolor{gray!30}\checkmark &  \cellcolor{gray!30}ADD $T$ & \cellcolor{gray!30}87.9 & \cellcolor{gray!30}57.2 $\pm$ \scriptsize 1.0 & \cellcolor{gray!30}\textbf{55.8} $\pm$ \scriptsize 5.9 & \cellcolor{gray!30}\textbf{57.1} $\pm$ \scriptsize 1.1 & \cellcolor{gray!30}\textbf{56.3} $\pm$ \scriptsize 3.2 \\
\\[-1.8ex]
\hline
\textcolor{blue}{\textit{Importance of prompts in POET}} & & & & & & & &  \\

POET w/o \textbf{Prompts} & \tikzxmark &  &  & 87.9 & 60.8 $\pm$ \scriptsize 0.5 \textcolor{blue}{(+3.6)} & 18.4 $\pm$ \scriptsize 1.0 \textcolor{red}{(-37.4)} & 57.3 $\pm$ \scriptsize 0.4 & 28.3 $\pm$ \scriptsize 1.1 \\
\hline
POET w/o \textbf{C.U.P.} & \checkmark & \checkmark & ADD $T$ & 89.2 & 45.5 $\pm$ \scriptsize 1.4 \textcolor{red}{(-11.7)} & 53.6 $\pm$ \scriptsize 3.7 \phantom{0}\textcolor{red}{(-2.2)} & 46.2 $\pm$ \scriptsize 1.2 & 49.1 $\pm$ \scriptsize 1.5 \\

POET w/o \textbf{\{Prompts, C.U.P.\}} & \tikzxmark & & & 88.4 & 40.0 $\pm$ \scriptsize 1.6 \textcolor{red}{(-17.2)} & 51.0 $\pm$ \scriptsize 2.3 \phantom{0}\textcolor{red}{(-4.8)} & 44.8 $\pm$ \scriptsize 1.1 & 40.9 $\pm$ \scriptsize 1.4 \\

POET w/o \textbf{\{Prompts, C.U.P., Freezing\} }& \tikzxmark & & & 88.4 & \phantom{0}0.2 $\pm$ \scriptsize 0.5 \textcolor{red}{(-57.0)} & 36.0 $\pm$ \scriptsize 10.1 \textcolor{red}{(-19.8)} & \phantom{0}3.2 $\pm$ \scriptsize 0.8 & \phantom{0}0.3 $\pm$ \scriptsize 1.0 \\ 
\hline
\textcolor{blue}{\textit{Impact of Ordered Selection}} & & & & & & & &  \\
POET w/o \textbf{Ordered Selection} & \checkmark & \checkmark & ADD $T$ & 88.2 & 59.9 $\pm$ \scriptsize 1.1 \textcolor{blue}{(+2.7)} & 52.5 $\pm$ \scriptsize 4.2 \phantom{0} \textcolor{red}{(-3.2)} & 59.3 $\pm$ \scriptsize 0.9 & 55.9 $\pm$ \scriptsize 2.2 \\
\hline
\textcolor{blue}{\textit{Relative importance of our Prompt Selection versus Prompt Integration}} & & & & & & & &  \\
POET Selection w/o \textbf{Addition}  & \checkmark & \checkmark & Cross-Attend $T$ & 82.9 & 57.0 $\pm$ \scriptsize 2.0 \textcolor{red}{(-0.2)} & 31.0 $\pm$ \scriptsize 4.8 \textcolor{red}{(-24.8)} & 54.9 $\pm$ \scriptsize 1.8 & 39.9 $\pm$ \scriptsize 4.0 \\

POET Addition w/o \textbf{Selection}  & \checkmark & Standalone & ADD $T$ & 88.6 & 58.8 $\pm$ \scriptsize 3.2 \textcolor{blue}{(+1.6)} & 54.0 $\pm$ \scriptsize 3.4 \phantom{0}\textcolor{red}{(-1.8)} & 58.4 $\pm$ \scriptsize 1.2 & 56.2 $\pm$ \scriptsize 2.2 \\

POET Addition w/o \textbf{Selection}  & \checkmark & Standalone & ADD $1$ & 88.2 & 56.9 $\pm$ \scriptsize 1.1 \textcolor{red}{(-0.3)} & 54.7 $\pm$ \scriptsize 5.3 \phantom{0}\textcolor{red}{(-1.1)} & 56.7 $\pm$ \scriptsize 1.2 & 55.7 $\pm$ \scriptsize 3.1 \\

POET w/o \textbf{\{Selection, Addition\}} & \checkmark & Standalone  & Cross-Attend $T$ &  43.3 & 25.6 $\pm$ \scriptsize 1.3 \textcolor{red}{(-31.6)} & \phantom{0}5.0 $\pm$ \scriptsize 4.1 \textcolor{red}{(-50.8)} & 23.9 $\pm$ \scriptsize 1.2 & \phantom{0}7.8 $\pm$ \scriptsize 4.9 \\
\bottomrule
\end{tabular}%
}
\label{tab:10 runs averaged}
\vspace{-10pt}
\end{table}

\subsubsection{How Well Does POET Mitigate Catastrophic Forgetting?}
In this section we dive deeper into understanding how our proposed Prompt Offset Tuning methodology \textit{mitigates catastrophic forgetting} in our few-shot class incremental setting for action recognition. We report mean and standard deviation across 10 experimental runs (10 sets of few-shots) for robustness ablation. The `Old' only accuracy gives a direct comparison of the forgetting. 

Our classifier expands in each continual session and fine-tuning the entire classifier on cross entropy loss of new classes leads to erosion of previous class weights in the classifier. \textbf{`C.U.P.'} refers to our classifier update protocol used to prevent forgetting from the classifier. We use a cosine normalized classifier for NTU RGB+D dataset (see Sec. \ref{setion: classifier update protocol} which helps mitigate forgetting from the classifier as shown by the $11.7\% \downarrow$ in `Old' class performance of \textbf{`POET w/o C.U.P.'}, w.r.t. POET.

In \textbf{`POET w/o Prompts'}, we simply remove our prompts altogether, keeping the cosine normalized classifier to observe the prompt only affect. While Old performance is slightly better, the backbone doesn't learn new knowledge as shown by $37.4\% \downarrow$ in `New' class performance. If we remove both prompts and classifier update protocol \textbf{`POET w/o Prompts, C.U.P.'}, we get vanilla Feature Extraction without any freezing or regularizing of classifier weights. It suffers in both Old and New classes. Further, as highlighted in the main paper as well, freezing our backbone during the continual user sessions $t>0$ is of prime importance given our few-shot setting where the overfitting to few training samples exacerbates the overwriting of existing knowledge, leading to a complete washout of previous knowledge as seen by the 0.2 Old performance.   

Notice, in the absence of our ordered prompt selection, New performance suffers by $3.3\%$ as the same prompts are selected and updated everytime. In POET, we ensure the prompts get selected in the right temporal sequence, learning new temporal semantics for new classes hence enabling better adaptation to new classes.

Finally, within prompts, we replace our prompt selection mechanism completely by standalone $T$ prompts or a single prompt. This shows how POET's spatio-temporal temporally consistent selection mechanism helps learn new orderings of T prompts as compared to attaching prompts without selecting them using an input dependent query. We also use cross attention along the temporal dimension as attachment operator $f_p(.)$ and find addition consistently outperforms. \textbf{`POET w/o {Selection, Addition}'} experiment shows the importance of our design choices as unsuitable selection and integration mechanisms can be fatal to continual learning performance.

\begin{table}[t]
\centering
\caption{\textbf{Backward Forgetting Metric (\%, $\downarrow$):} Here we present BWF after user session \task{4}. POET significantly mitigates forgetting on old classes.}
\resizebox{0.35\textwidth}{!}{%
\setlength{\tabcolsep}{2pt}
\begin{tabular}{@{}l||c@{}} \toprule
         & \normalsize \tabletask{4} \\ 
\textbf{Method} & Forgetting ($\downarrow$) \\
\hline
\\[-1.8ex]
 \cellcolor{orange!30}FE+Replay & \cellcolor{orange!30} -0.90 $\pm$ \scriptsize 0.65 \\
\hline
FE &  52.83 $\pm$ \scriptsize 2.09 \\
FT & 54.08 $\pm$ \scriptsize 9.17 \\
FE, Freeze &  34.02 $\pm$ \scriptsize 0.83 \\
\hline
\hline
 \cellcolor{gray!30}\textbf{POET (Ours)} & \cellcolor{gray!30} 29.91 $\pm$ \scriptsize 0.34  \\
\bottomrule
\end{tabular}%
}
\label{BWF}
\vspace{-10pt}
\end{table}

\subsubsection{Backward Forgetting Metric (BWF):}

In addition to these results, we report the average forgetting metric \cite{chaudhry2018riemannian, chaudhry2018efficient} after the model has been trained continually on all user sessions (after \task{4}) as: 
\begin{equation}
F_k = \frac{1}{k-1} \sum_{j=1}^{k-1} f_j^k
\end{equation}

where $f_j^k$ is the forgetting on previous task `j' after the model is trained with all the few-shots up till task $k$:

\begin{equation}
f_j^k = \max_{l \in \{1, \ldots, k-1\}} a_{l, B_{l,j}} - a_{k, B_{k,j}}
\end{equation}

\noindent In effect, this is the same as difference in performance of each previous task `j' at the end of \task{4} from when the task was first introduced (\textit{New} in \task{j}).  
\subsubsection{Role of number of few-shots in continual learning:}
\begin{figure*}[h!]
     \centering
     \begin{subfigure}[b]{0.45\textwidth}
         \centering
         \includegraphics[width=\textwidth]{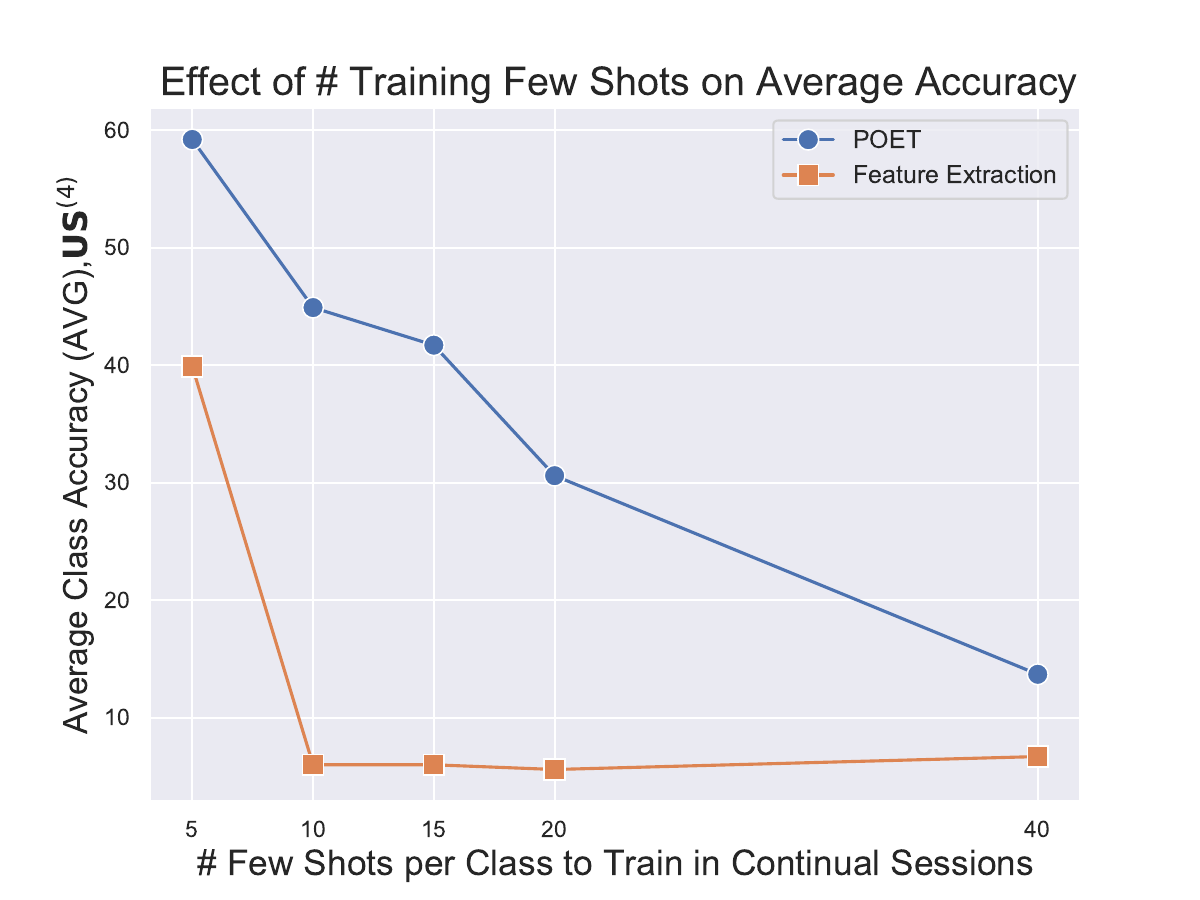}
         \caption{Average Accuracy (AVG)}
         \label{fig: AVG}
     \end{subfigure}
     \hfill
     \begin{subfigure}[b]{0.45\textwidth}
         \centering
         \includegraphics[width=\textwidth]{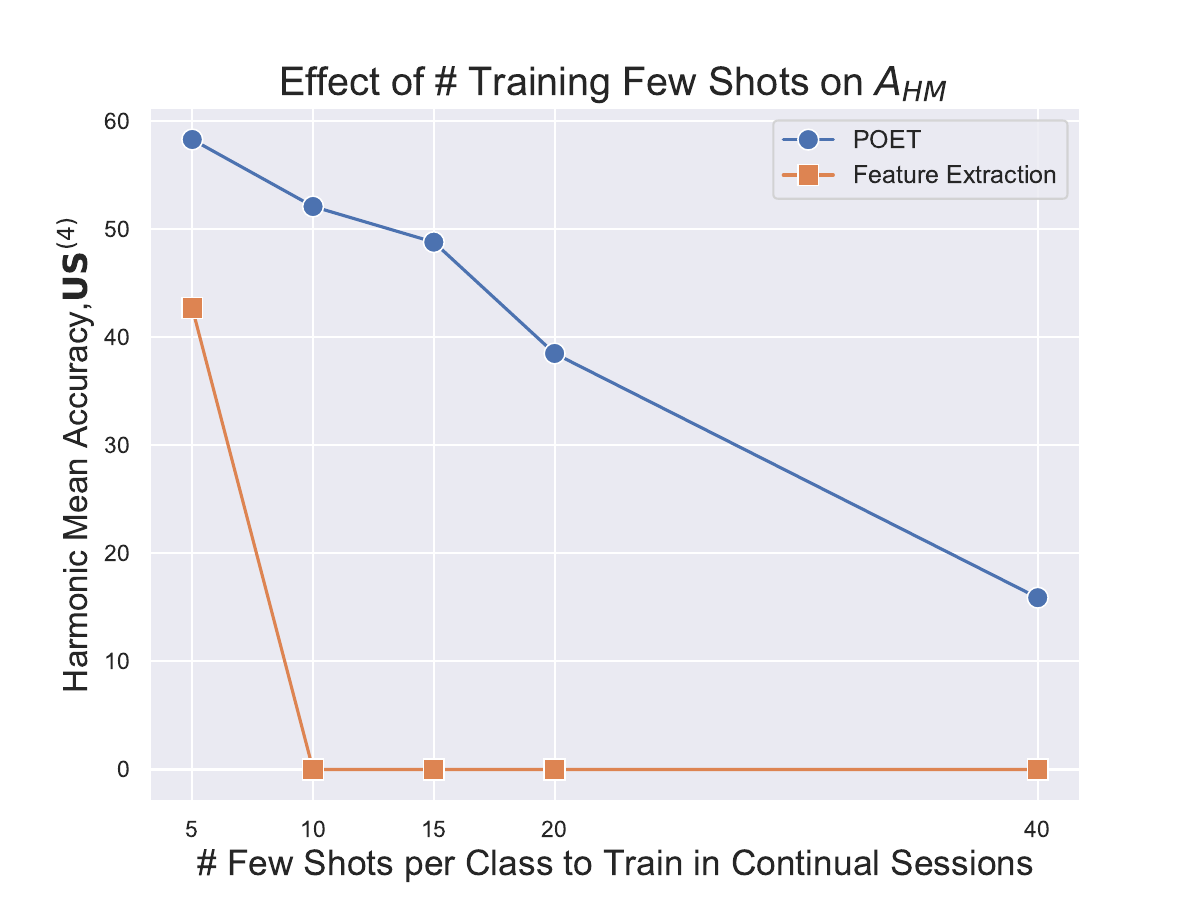}
         \caption{$A_{HM}$}
         \label{fig: AHM}
     \end{subfigure}
     \caption{Effect of variation in number of few-shot samples used for training in user sessions \task{1}-\task{4} on stability-plasticity trade-offs in our few-shot continual setting.}
     \label{fig: vary FS}
\end{figure*}

We also vary the number of few shots used for training per new class in continual sessions in Fig \ref{fig: vary FS}. The Feature Extraction baseline reduces to zero $A_{HM}$ because the Old class performance reduces to zero. We find our prompt offsets in POET (=FE+Prompts) significantly help retain old class performance as compared to Feature Extraction, without any explicit forgetting measure due to our ordered prompt selection and clustering loss. It can be noted that our method is particularly well suited for very few training samples per user (<15) and may require additional explicit regularization or freezing of prompt pool to mitigate forgetting for large number of training samples (>20).

\subsection{Impact of Prompts in POET}
\begin{figure*}[h]
\centering
\includegraphics[width=0.5\textwidth]{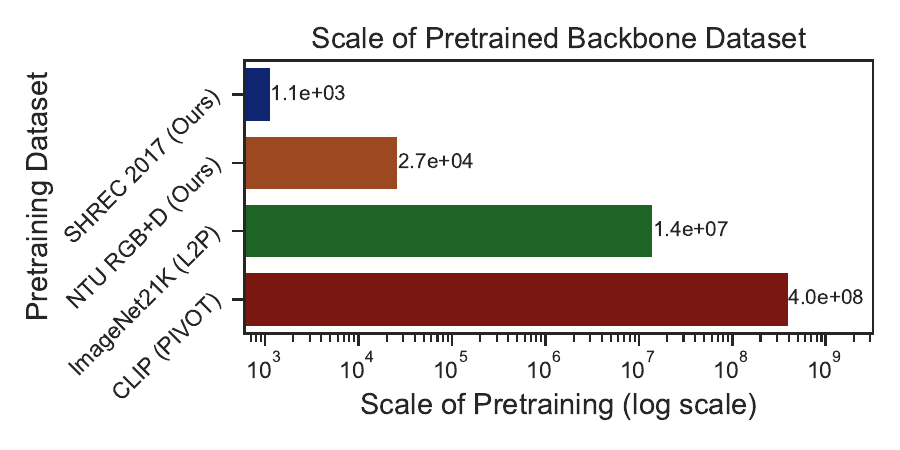}
\caption{\footnotesize \textbf{Scale of pretraining used for the prompt tuning backbones.} For (Our) benchmarks on NTU RGB+D and SHREC 2017, numbers represent the base class training data used. Our POETs continually learn new actions mitigating catastrophic forgetting, without massive pretraining, and only rely on prompts.}
\label{fig: scale}
\end{figure*}

In Figures \ref{fig: power of prompts, NTU} and \ref{fig: power of prompts, SHREC}, we qualitatively study the impact of prompts by removing prompts from POET (the corresponding feature extraction baselines `FE' for NTU RGB+D and `FE++' for SHREC). Prior continual learning works rely on ImageNet21K pretrained ViT \cite{ridnik2021imagenet} ( L2P \cite{wang2022learning}, Dual-P \cite{wang2022dualprompt}, CODA-P \cite{smith2023coda}) or WebImageText pretrained CLIP model \cite{radford2021learning} (S-Prompts \cite{wang2022s}, PIVOT \cite{villa2023pivot}) for prompt tuning. In Fig. \ref{fig: scale}, we show the significant disparity in scale of pretraining dataset as we use only base class dataset from the benchmark itself for pretraining and every new user session sees a non-overlapping set of classes. Despite of this, POET shows promising results. 

\begin{figure*}[t!ht]
\centering
\includegraphics[width=0.99\textwidth]{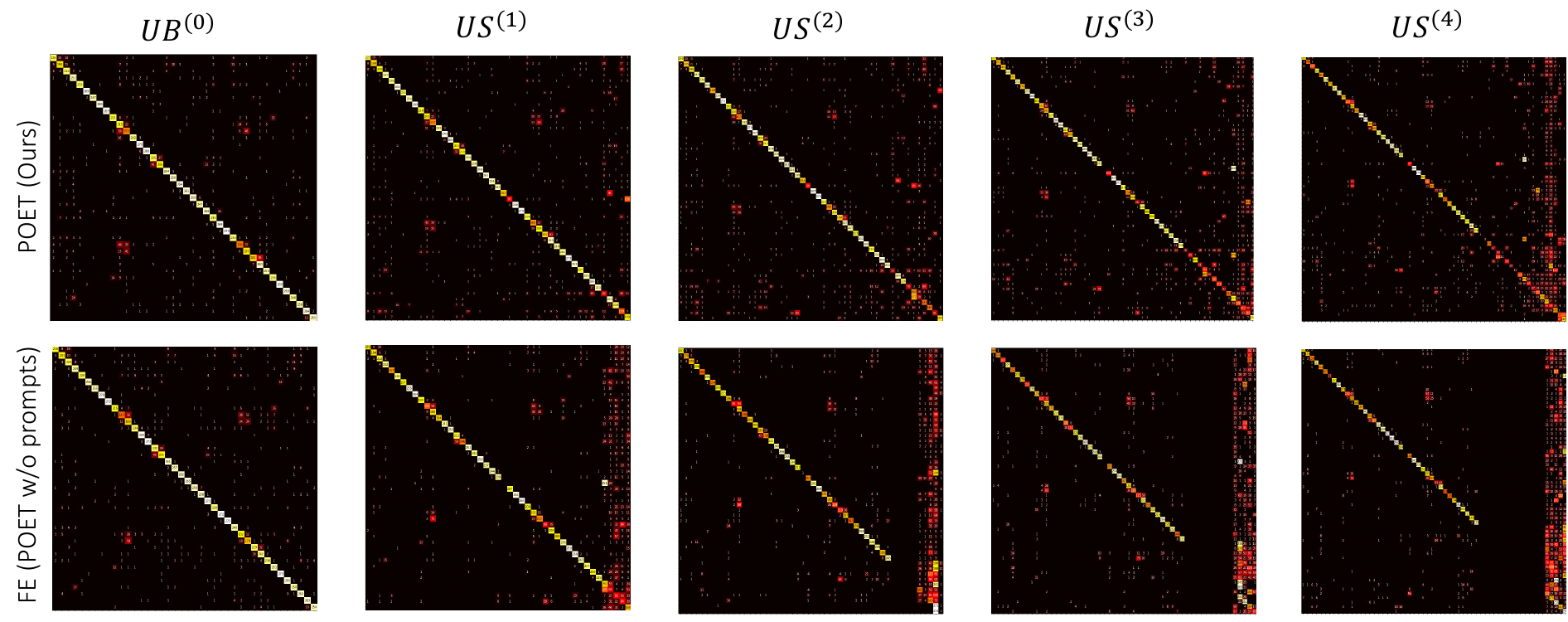}
\caption{\textit{Confusion matrices showing the impact of our prompt offsets across 4 user sessions in \textbf{NTU RGB+D activity recognition benchmark}}. We compare confusion across new and old actions in POET and POET w/o prompts ablation. Starting from 40 default classes in $\mathcal{UB^{(0)}}$, we learn new classes \textcolor{violet}{\{\task{1}: sneeze, stagger, fall, touch head, touch chest\}}; \textcolor{violet}{\{\task{2}: touch back, touch neck, nausea, user fan, punch\}}; \textcolor{violet}{\{\task{3}: kick, push, pat back, point finger, hug\}}; \textcolor{violet}{\{\task{4}: give, touch pocket, handshake, walk towards, walk away\}}. Prompts enable retention of the intermediate \textbf{`New-Old'} classes very well, while FE gets heavily biased towards the new classes (see last 5 columns in each matrix).}
\label{fig: power of prompts, NTU}
\end{figure*}
\vspace{-0.1cm}

\begin{figure*}[h!b]
\centering
\includegraphics[width=0.8\textwidth]{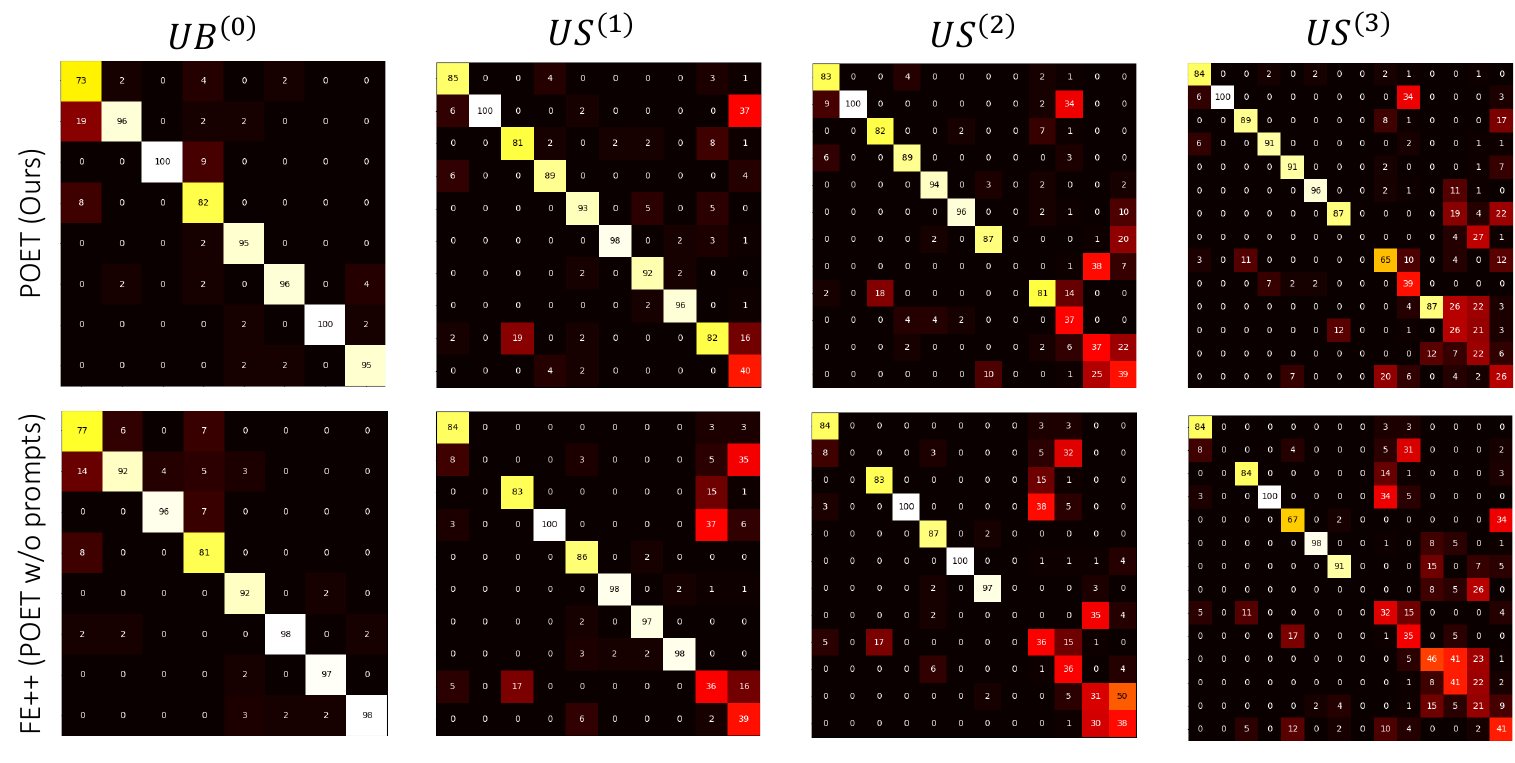}
\caption{\textit{Confusion matrices showing the impact of our prompt offsets across 3 user sessions in \textbf{SHREC 2017 gesture recognition benchmark}}. \task{0} has hand gestures \textcolor{violet}{grab, tap, expand, pinch, rotate clockwise, rotate counter-clockwise, swipe right, swipe left\}}. \textcolor{violet}{\{\task{1}: swipe up, swipe down\}}, \textcolor{violet}{\{\task{1}: swipe-x, swipe-+\}}, \textcolor{violet}{\{\task{1}: swipe-v, shake\}}. Even though the classes are fine-grained, the prompts help retain old class semantics well.}
\label{fig: power of prompts, SHREC}
\end{figure*}

\begin{figure*}[t]
    \centering
    \begin{subfigure}[t]{0.5\textwidth}
        \centering
        \includegraphics[height=1.4in]{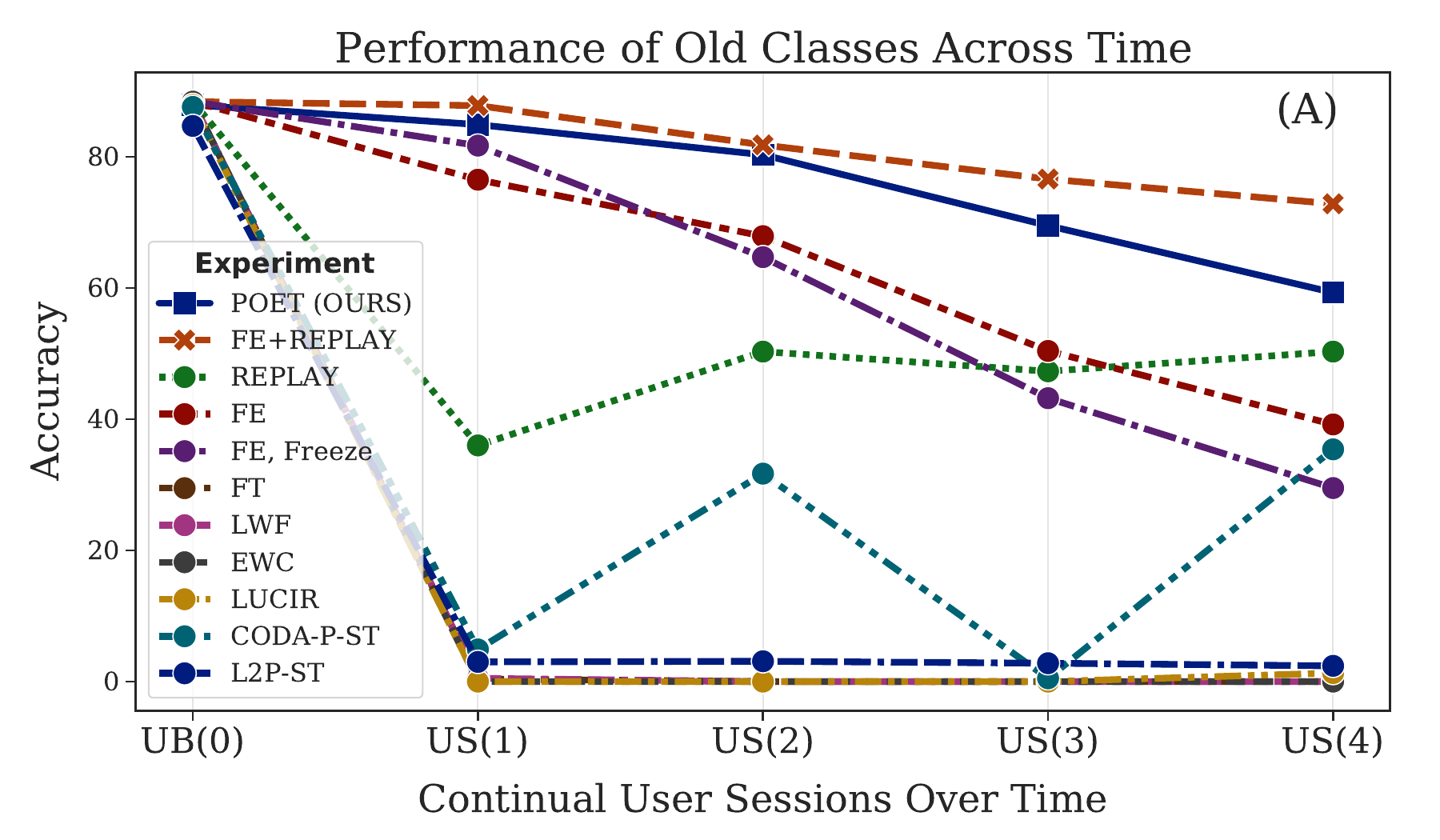}
    \end{subfigure}%
    ~ 
    \begin{subfigure}[t]{0.5\textwidth}
        \centering
        \includegraphics[height=1.4in]{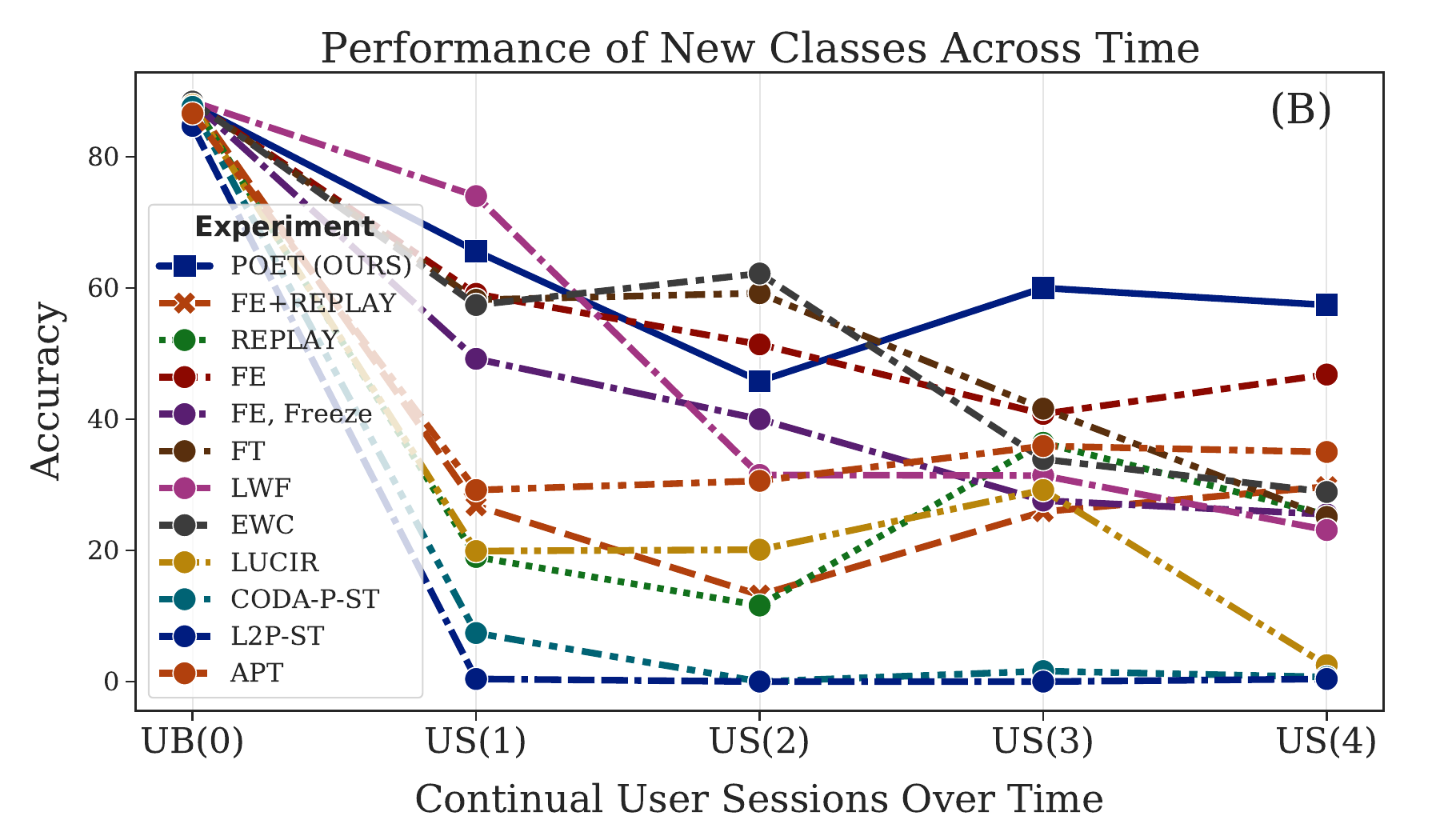}
    \end{subfigure}
    \caption{Old and New class performance for NTU RGB+D.}
    \label{fig: old-new-ntu}
\end{figure*}

\begin{figure*}[b!ht]
    \centering
    \begin{subfigure}[t]{0.5\textwidth}
        \centering
        \includegraphics[height=1.35in]{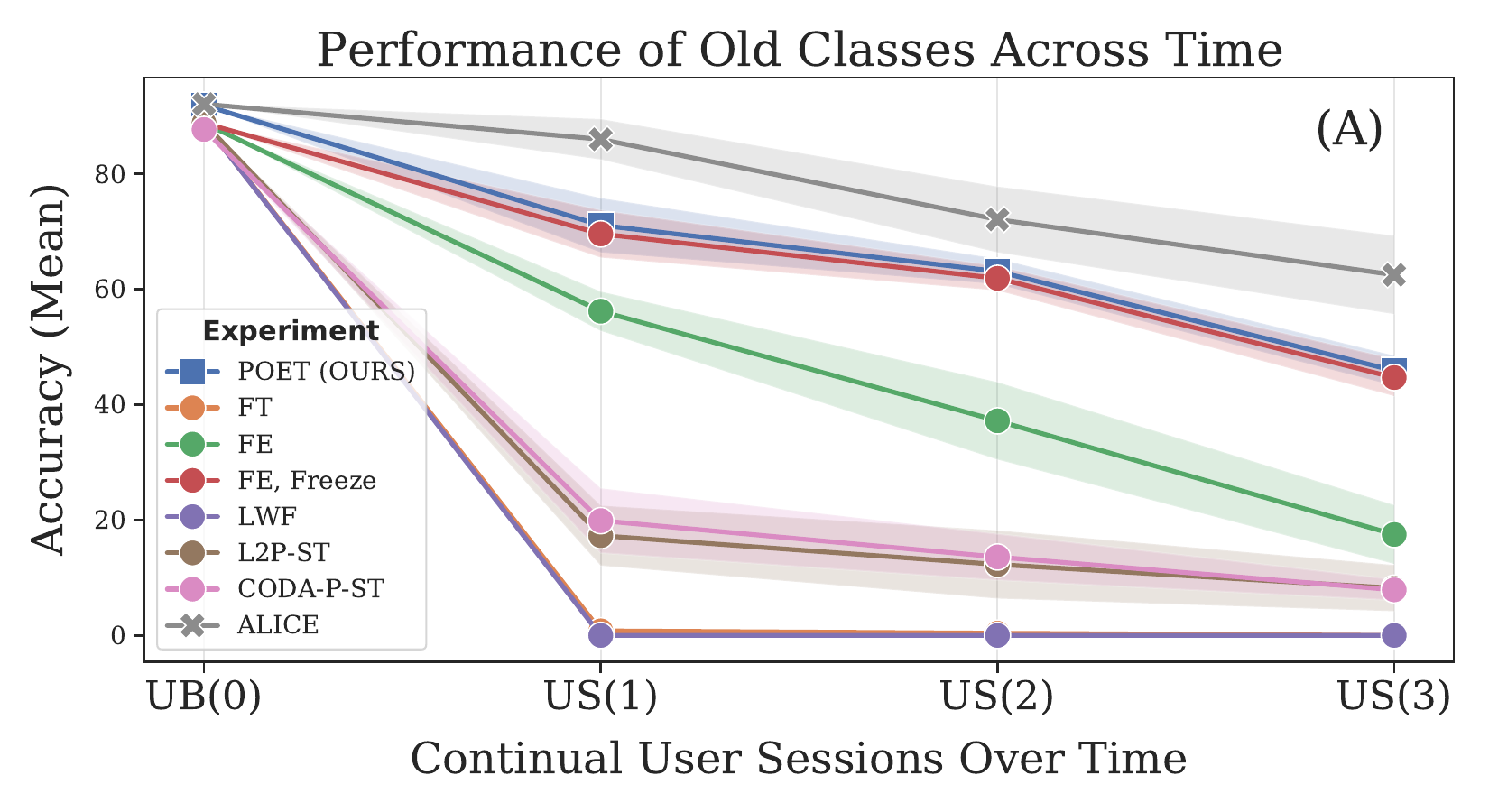}
    \end{subfigure}%
    ~ 
    \begin{subfigure}[t]{0.5\textwidth}
        \centering
        \includegraphics[height=1.35in]{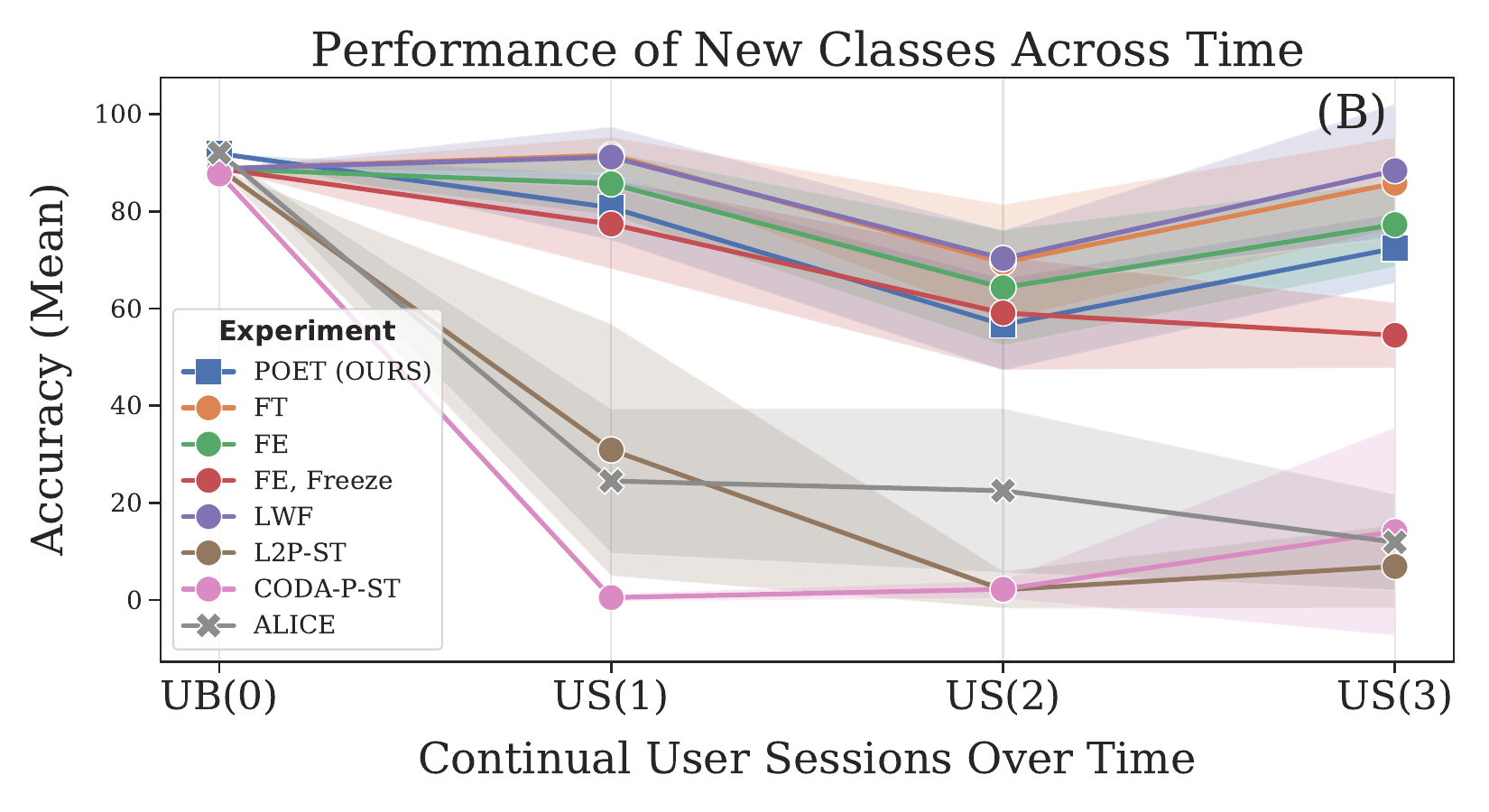}
    \end{subfigure}
    \caption{Old and New class performance for SHREC 2017. Reporting Mean and STD over 5-sets of user few-shots.}
    \label{fig: old-new-shrec}
\end{figure*}

\subsection{Stability-Plasticity Trade-offs via New/Old performance}
In Fig. \ref{fig: old-new-ntu}, we study the average accuracy of only new classes (New) and only old classes (Old) after every user session in activity recognition benchmark. As stated in Sec \ref{experiment section} of main paper, we observe that (i) any method that does not freeze the backbone such as knowledge distillation (LWF and LUCIR), prior-based regularization methods (like EWC), or vanilla Fine-tuning baselines (FT) completely forget old performance from \task{1} itself. POET is short of only FE+Replay which is an upper bound. (ii) Any existing prompt tuning work which does not update their query function such as L2P, CODA-P, APT in graph (B), is unable to learn new classes well.

In Fig. \ref{fig: old-new-shrec}, we observe similar trends. Additionally, (i) ALICE retains old knowledge very well as it does not use a parametric classifier for incremental sessions. However, ALICE is unable to learn new classes well. We also observe that (ii) DG-STA backbone has very high plasticity when fine-tuned on new tasks (see New performance of FT and LWF in graph B). But these baselines while plastic, completely forget old classes. POET achieves the best stability-plasticity trade-offs (indicated by $A_{HM}$ in main paper).

\subsection{Robustness to class order in user sessions}
\begin{figure*}[th]
\centering
\includegraphics[width=0.5\textwidth]{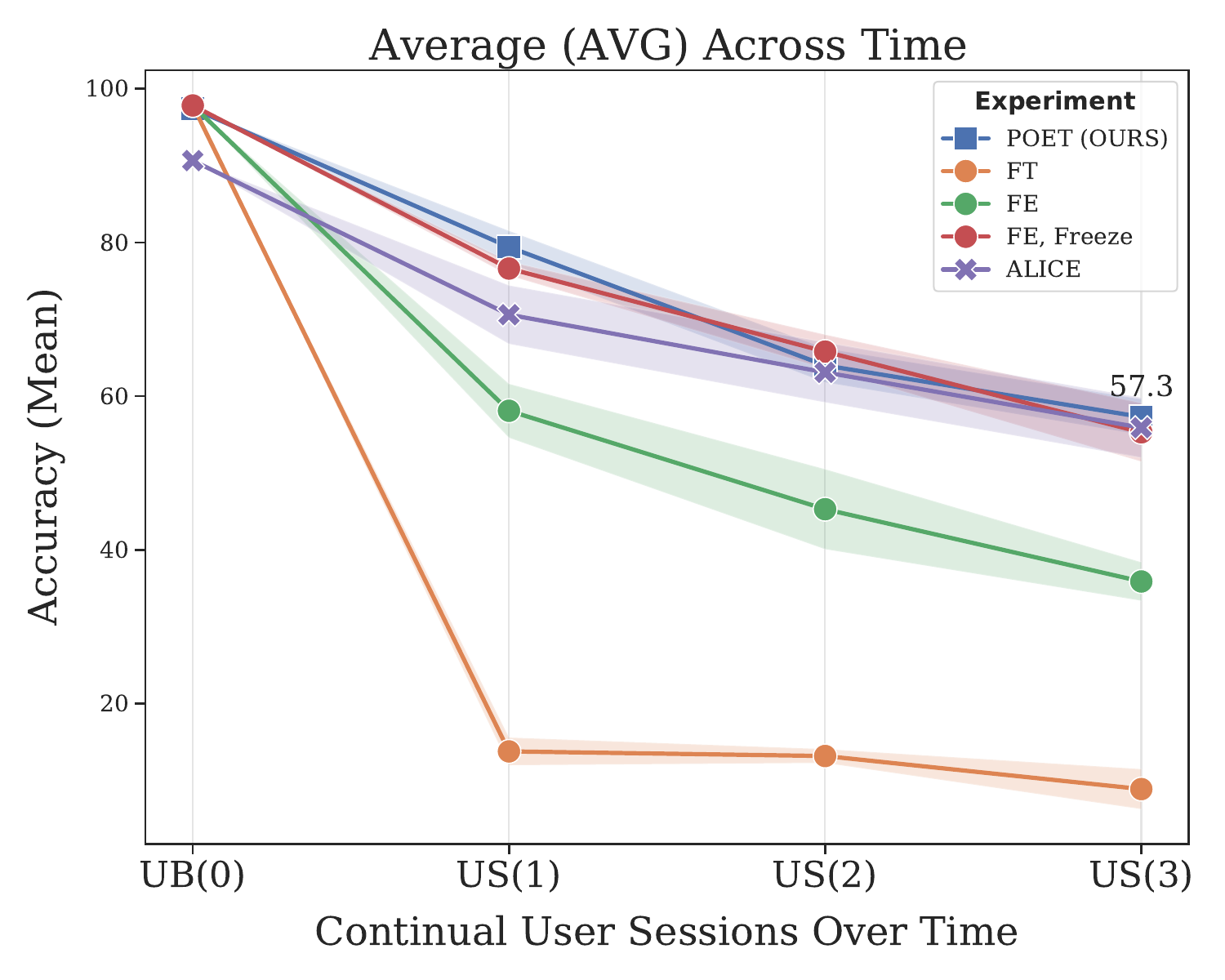}
\caption{Here, we change the set of classes in each session from the default order seen before. We report average accuracy of all classes learnt by the model after adding each new session. New ordering: \{\task{0}: Swipe-R, Swipe-L, Swipe-U, Swipe-D, Swipe-x, Swipe-+, Swipe-v, Shake\} $\rightarrow$ \{\task{1}: Grap, Tap\} $\rightarrow$ \{\task{2}: Expand, Pinch\} $\rightarrow$ \{\task{3}: Rotate-CW, Rotate-CCW\}.}
\label{fig: clsorder}
\end{figure*}

The default continual order in which different gesture classes appear till now was: \{\task{0}: Grab, Tap, Expand, Pinch, Rotate-CW, Rotate-CCW, Swipe-R, Swipe-L\} $\rightarrow$ \{\task{1}: Swipe-U, Swipe-D\} $\rightarrow$ \{\task{2}: Swipe-x, Swipe-+\} $\rightarrow$ \{\task{3}: Swipe-v, Shake\}. In Fig \ref{fig: clsorder}, we swap the base and incremental classes in SHREC benchmark to a new ordering: \{\task{0}: Swipe-R, Swipe-L, Swipe-U, Swipe-D, Swipe-x, Swipe-+, Swipe-v, Shake\} $\rightarrow$ \{\task{1}: Grap, Tap\} $\rightarrow$ \{\task{2}: Expand, Pinch\} $\rightarrow$ \{\task{3}: Rotate-CW, Rotate-CCW\}. We find \textbf{`POET'} gives an $AVG=57.3$ as compared to \textbf{`ALICE'}, $AVG=55.9$ and \textbf{`FE, Freeze'}, $AVG=55.3$ at the end of 3 user sessions even though our backbone is now trained on a different set of classes and we completely reversed the semantic order in which prompts learn different fine-grained gesture classes. This demonstrates robustness to variation in continual class order across tasks. 

\subsection{Ordered Key Index Selection $(s_i)_{i=1}^{T}$: Qualitative Results}
\begin{figure*}[thp]
\centering
\includegraphics[width=0.97\textwidth]{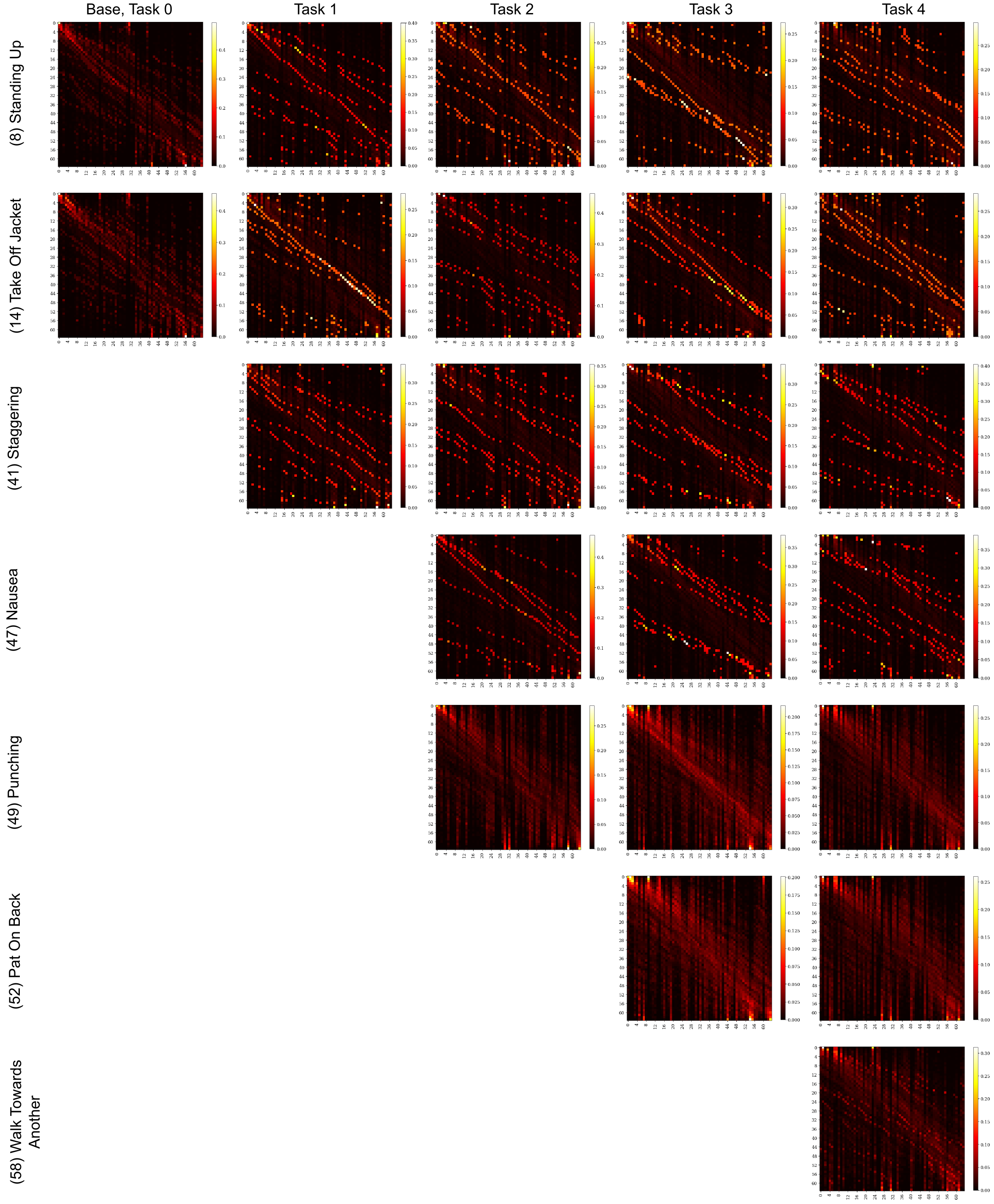}
\caption{\textbf{Class-level Ordered Prompt Selection using POET, Task t is same as \task{t}:} Here, we analyse the ordered prompt selection statistics for our method for different classes at \textbf{test time}. For each class shown in column 1, we plot the prompt selection order at test time for each continual model checkpoint (starting from when that class was first introduced to the continual system and checking after updating the model on new classes each time). We \textbf{observe} that class-wise selection statistics are retained even after Task 4 (notice the plots for different classes in Task 4). Even for classes introduced as part of the same task (class 47, Nausea and class 49, Punching both introduced in Task 2), their ordered prompt selection is unique and consistent even after updating the model on new data in subsequent continual sessions. }
\label{fig: class-level order matrices}
\end{figure*}

\begin{figure*}[thp]
\centering
\includegraphics[width=0.9\textwidth]{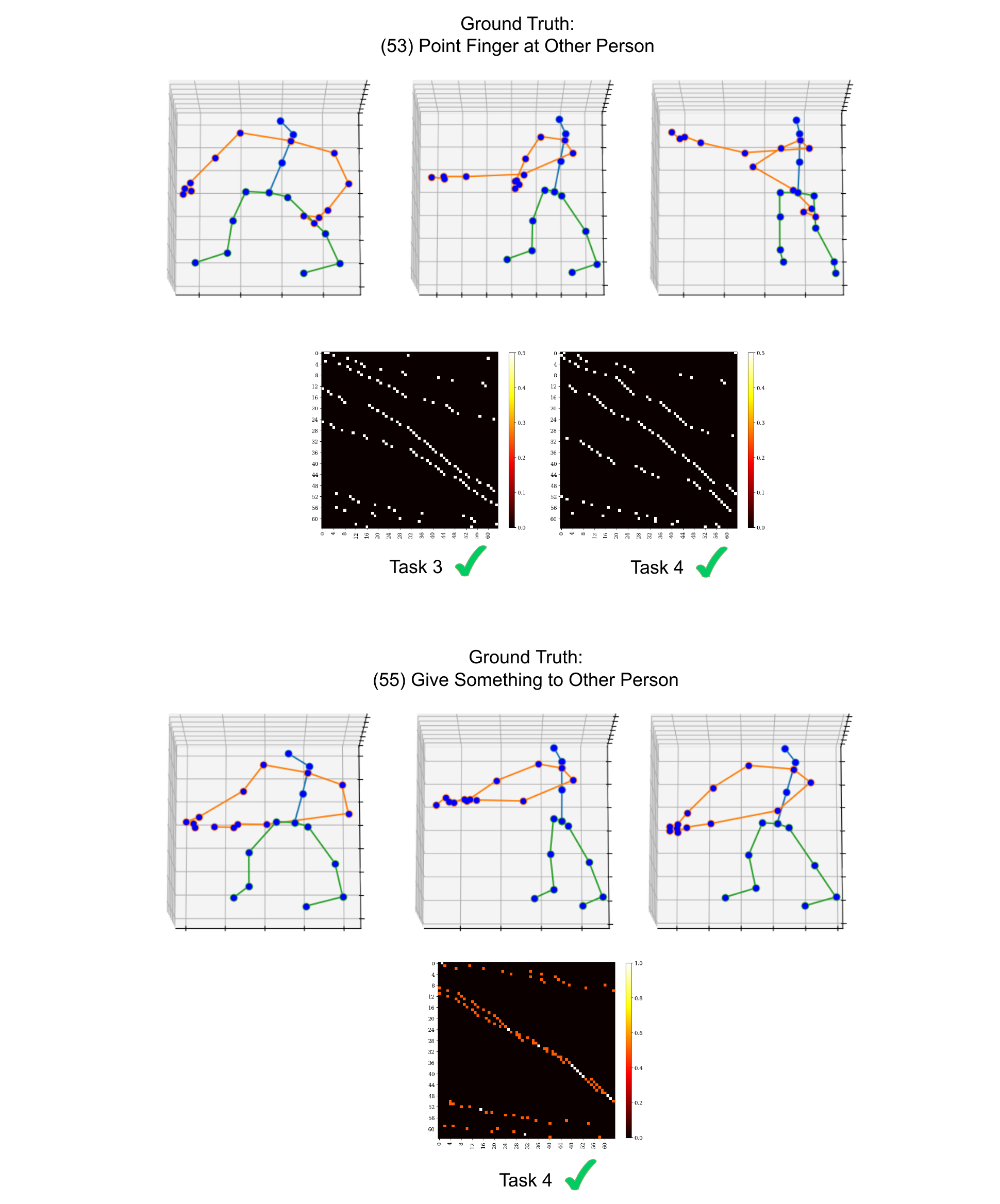}
\caption{\textbf{Instance-Level Ordered Prompt Selection using POET:} Our proposed method POET is an input instance-based prompt tuning approach for FSCIL, as the prompts are selected conditioned on each input instance itself. Hence, here we study instance-level prediction on the test set. The sample of class \textit{Point Finger, class ID 53} is evaluated after \task{3} and \task{4} as the class was added to the model in \task{3}. The sample of class \textit{Give Something, class ID 55} is continually learnt and evaluated after \task{4}. We point out the unique ordered key index sequence for the 2 instances, which could have been easily confused by the model due to their semantic similarity. The ordering matrix for \textit{Point Finger} remains consistent across tasks, even after adding 5 new classes in \task{4}.}
\label{fig: correct sample}
\end{figure*}

\begin{figure*}[thp]
\centering
\includegraphics[width=0.99\textwidth]{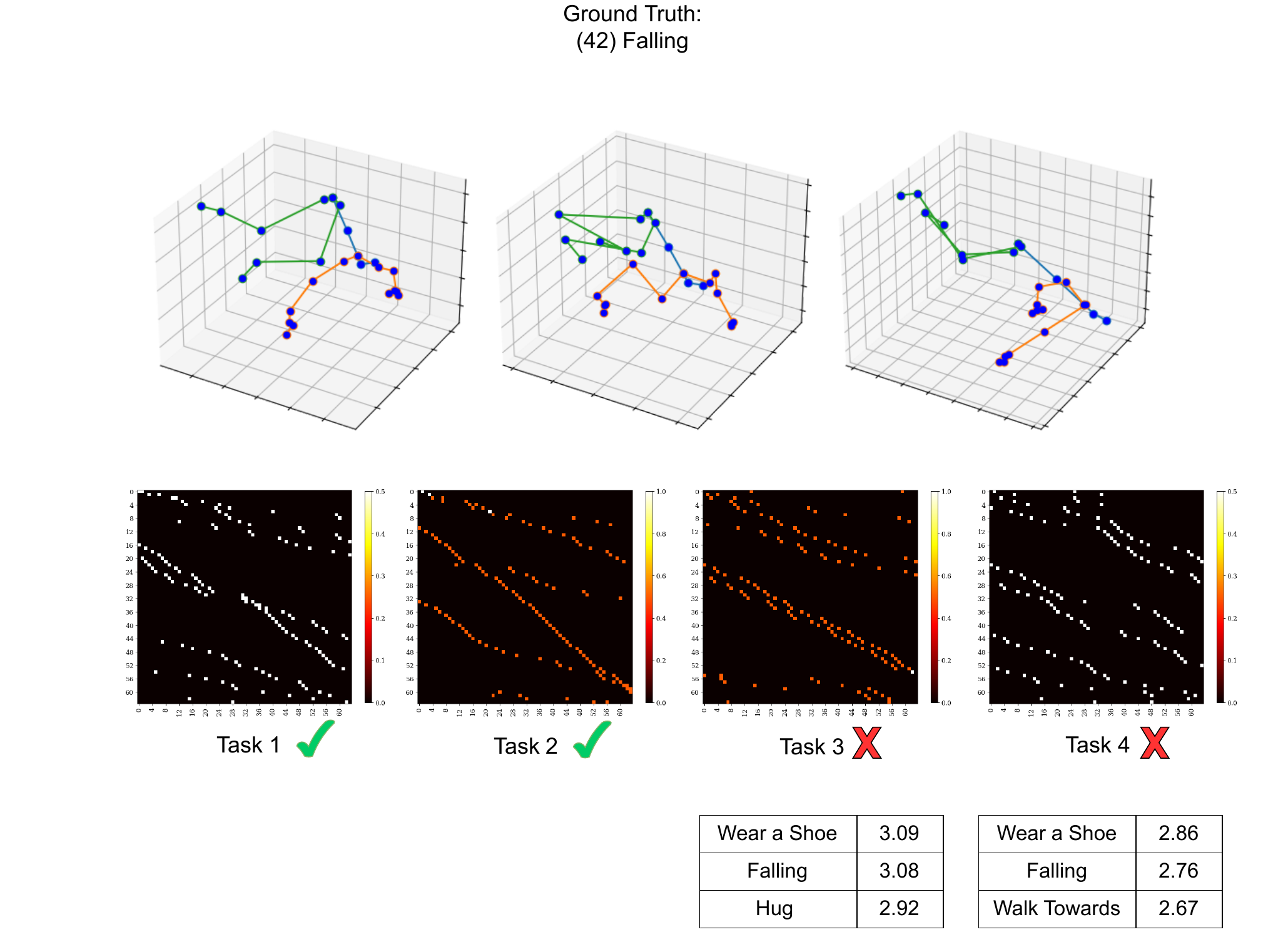}
\caption{\textbf{Instance-Level Ordered Prompt Selection using POET:} We also show a failure case of our proposed approach. After learning the class \textit{Falling} in \task{1}, we evaluate it after every new continual task. Even though it correctly predicts a test set instance in \task{1} and \task{2}, it tends to get confused by the class \textit{Wearing a Shoe} at  \task{3} and \task{4}. Notice, this coincides with a disruption in the ordering statistics.}
\label{fig: incorrect sample}
\end{figure*}

In Section \ref{sec: method} of main paper, we explained our sorted \textit{ordered key index selection} for selecting temporally consistent prompts from the pool. In Figure \ref{fig: method, prompt selection} of main paper, we visualized $(s_i)_{i=1}^{T}$ ordering statistics at the task-level. In Figure \ref{fig: class-level order matrices}, we investigate class-wise ordering statistics at inference time, performing inference after each continual task. We find that the ordering statistics are not only disparate for different classes, the statistics for a class remain consistent across continual tasks. The temporal discriminability in these studies further establishes that our learnable prompt selection mechanism is temporally with the 4D input skeleton. Finally, we also demonstrate instance-level statistics in Figure \ref{fig: correct sample} and Figure \ref{fig: incorrect sample} as our prompt mechanism is designed to select relevant (temporal ordered) prompts conditioned on every input instance. This means that our method does not depend on disparate task-wise or class-wise dataset splits and can even be used for online continual learning settings that do not have clear task boundaries. 

\subsection{Prompt Pool Expansion}
We further present the order-preserving prompt pool expansion Algorithm \ref{alg:expansion} that (A) expands pool to learn new knowledge, (B) freezes previous prompts to prevent forgetting, and (C) forces usage of new prompts at the end of the sequence, hence alleviating the prompt pool collapse issue while preserving already existing temporal order statistics Figure \ref{fig: pool collapse} of main paper. We find $R=6$ new prompts to be the best empirically for NTU RGB+D and $R=2$ for SHREC 2017 using our 30\% validation set of incremental sessions. Algorithm \ref{alg:expansion} presents our algorithm for prompt pool expansion. 
\begin{algorithm*}[h]
\small
\caption{Prompt Pool Expansion at Train Time, $t \geq 1$}\label{alg:expansion}
\begin{algorithmic}
\State \textbf{Input:} Query function $f_q$, keys $\boldsymbol{K}=\{\boldsymbol{k_j}\}_{j=1}^T$, prompt pool $\mathbf{P}=\{\boldsymbol{P_j}\}_{j=1}^T$; main model $f_e$, $f_g$, $f_c$
\State \textbf{Expand:} \\ Pool and keys by $R$ new prompts as: $\mathbf{P}_M \rightarrow \mathbf{P}_{M+R} ; \boldsymbol{K}_M \rightarrow \boldsymbol{K}_{M+R}$
\State Where $\mathbf{P}_{M+R} = \{\mathbf{P}_M;\mathbf{P}_R\}$ (attach new prompts at the end of existing tensor)
\State \textbf{Initialize:} New prompts $\mathbf{P}_i \leftarrow \mathcal{U}(0,1)$; new keys $\boldsymbol{K}_i \leftarrow Mean(\boldsymbol{K}_M)$
\State \textbf{Construct} $\mathbf{P}_{T}$ as: 
    \State \hskip0.8em 1. Find $T - R$ key indices $\boldsymbol{K}_{T-R}$ using Eq. \ref{eqn: 7}. Use this sequence to read previous prompts in the pool $\mathbf{P}_M$ and form $\mathbf{P}_{T-R}$.
    \State \hskip0.8em 2. Concatenate $\mathbf{P}_R$ new prompts at the end of the sequence: $\mathbf{P}_{T} = \{\mathbf{P}_{T-R} ; \mathbf{P}_R\}$ (i.e.\ explicitly use $R$ new prompts).
\State \textbf{Freeze:} Previous task prompts in the pool \textcolor{cyan}{$\mathbf{P}_M$}. 
\State \textbf{Train:} New prompts \textcolor{RedOrange}{$\mathbf{P}_R$}, all keys \textcolor{RedOrange}{$\boldsymbol{K}_{M+R}$} (to learn global inter-task selection), query adaptor \textcolor{RedOrange}{$f_{QA}$}, and classifier \textcolor{RedOrange}{$f_c$}.
\end{algorithmic}
\end{algorithm*}

\section{Broader Impact and Limitations}

\textbf{Privacy-aware human action recognition in extended reality devices:} In order to protect users' privacy and security in head mounted devices, we incorporate privacy awareness in our continual learning setup: (i) By not storing any old class exemplar data or prototypes for replay in continual user sessions. All data is trained in a session and discarded henceforth. (ii) By using only 3D skeleton joint input modality for action recognition, we circumvent the visual privacy violation and user identity revelation in video-based HAR \cite{li2023stprivacy, hinojosa2022privhar, kumawat2022privacy}. While we are a privacy-aware continual learning setting, we do not claim differential privacy and studying differential privacy in our prompts will be an interesting future work direction.  

\noindent \textbf{Data-free adaptation of action models for new user categories:} Our key motivation for a \textit{data-free} prompt tuning-based action recognition model adaptation to new categories over time is to maintain privacy of past sessions' data. However, this also has other advantages. 
Firstly, a data-free solution does not require a memory budget on the edge device for replay of old class data. Secondly, it has become commonplace to have access to large pre-trained backbones, but there is limited knowledge and often lack of access to such a dataset for such pre-training. Finally, prompts via a vector-quantized prompt pool memory offer a compact, learnable and automatically retrievable bottleneck of task-specific information (like in auto-decoders). Even if businesses can store and retrain their model on all previous data to continually adapt to new data, training large models incurs high carbon footprints \cite{lacoste2019quantifying}. Prompts offer a cost-efficient and low carbon footprint solution  to retraining large models from scratch every time new data of value becomes available. Evaluating our design choices to large pre-trained models for skeletal data, as and when they become available, is another direction of future work.


\let\titleold\title
\renewcommand{\title}[1]{\titleold{#1}\newcommand{\thetitle}{#1}}
\def\maketitlesupplementary
   {
   \newpage
       \twocolumn[
        \centering
        \Large
        \textbf{\thetitle}\\
        \vspace{0.5em}Supplementary Material \\
        \vspace{1.0em}
       ] 
   }

\end{document}